\documentclass[journal]{IEEEtran}
\ifCLASSINFOpdf
\else
\fi

\usepackage{graphicx}
\usepackage{cite}
\usepackage{multirow}
\usepackage{verbatim}

\usepackage{url}
\usepackage{color,soul}

\usepackage{wrapfig}
\usepackage{mdframed}
\usepackage{fancybox}
\usepackage{xcolor}

\newcommand{\new}[2]{{#2}}

\begin{document}
%






\title{Deep Learning and Earth Observation to \\ Support the Sustainable Development Goals}

%
%
%

\author{Claudio Persello, Jan Dirk Wegner, Ronny Hänsch, Devis Tuia, Pedram Ghamisi, Mila Koeva \\ and Gustau Camps-Valls}


%
%

\markboth{IEEE Geoscience and Remote Sensing Magazine,~Vol.~X, No.~X, December ~2021}%
{Shell \MakeLowercase{\textit{et al.}}: Bare Demo of IEEEtran.cls for IEEE Journals}
%



\maketitle



%
\IEEEpeerreviewmaketitle

 
\emph{This is the pre-acceptance version of the paper. The final version will appear in the IEEE Geoscience and Remote Sensing Magazine.} 

{\bf
The synergistic combination of deep learning models and Earth observation promises significant advances to support the sustainable development goals (SDGs). New developments and a plethora of applications are already changing the way humanity will face the living planet challenges. This paper reviews current deep learning approaches for Earth observation data, along with their application towards monitoring and achieving the SDGs most impacted by the \new{}{rapid development} of deep learning in Earth observation.
We systematically review case studies to 1) achieve zero hunger, 2) sustainable cities, 3) deliver tenure security, 4) mitigate and adapt to climate change, and 5) preserve biodiversity.
Important societal, economic and environmental implications are concerned. Exciting times ahead are coming where algorithms and Earth data can help in our endeavor to address the climate crisis and support more sustainable development.
}

\section{Introduction}

Machine learning has played a fundamental role in the analysis of Earth observation (EO) data for over three decades, and its importance has been continuously growing. From the early investigations in artificial neural networks and statistical techniques \cite{Benediktsson1990NeuralData, Lee1990AClassification}, the EO community has been striving for effective algorithms to automate the extraction of information from various sources of remotely sensed images, in situ data and models. The developments in sensor technologies and the increasing availability of voluminous data go hand-in-hand with the demand for more accurate and scalable information extraction methods and tools. This demand is spurred by many geospatial applications and the growing awareness of the necessity to monitor system Earth for the multiple threats to our natural environment, climate, and the sustainable development of human societies.

After a long period when neural networks fell out of fashion, the deep learning (DL) revolution started about a decade ago, and brought back the attention to these powerful learning algorithms \cite{Krizhevsky2012ImageNetNetworks, Reichstein2019}. Thanks to the development of specialized hardware, i.e., graphical processing units (GPUs), and the availability of large benchmark data sets, DL networks became more and more popular. They revealed extremely versatile learning machines able to learn virtually any task in data and image analysis \cite{LeCun2015DeepLearning}. Neural networks can be seen as trainable data-processing graphs, where the input data is gradually transformed through a sequence of layers that extract intermediate features and are finally used to predict the target output. In a supervised setting, the network is trained with a set of input-output instances, exemplifying the functional relationship between the explanatory covariates and the target variable to predict. 
Although simplistic, this view depicts the flexibility of neural networks for data analysis purposes. A variety of architectures have been developed so far, the most popular being multilayer perceptrons (MLPs) \cite{Rumelhart1986LearningErrors}, convolutional neural networks (CNNs) \cite{LeCun1989BackpropagationRecognition}, recurrent neural networks (RNNs) \cite{Hochreiter1997}, autoencoders \cite{Hinton2006ReducingNetworks}, and generative adversarial networks (GANs) \cite{NIPS2014_5423}. Moreover, as the number of consecutive layers increases, i.e., the network becomes deeper, the algorithm tends to improve its ability to learn informative features capturing intricate structures within input variables and their relation with the target output.

In the context of EO applications, deep networks can address a large variety of analysis tasks, from image classification and segmentation to data fusion, change detection, object detection and delineation. Deep networks can be designed according to the characteristics of the remotely sensed data and possibly fuse different sensor data types and information layers. One of the main advantages of DL is the ability to learn abstract hierarchical representations of the data, allowing networks to bring  spatial, spectral, and temporal patterns hidden in the data to the surface. This results in state-of-the-art performance and enables researchers and engineers to streamline the information extraction processing chain, potentially integrating multi-modal data fusion, feature extraction, and inference tasks into one, holistic, end-to-end learning framework. The combination of DL with powerful computing infrastructure and massive EO data sets opens up tremendous opportunities for geospatial applications. 
State of the art DL methods are closing the gap between the performance of automated workflows and the need for accurate and reliable information imposed by real applications. 
\begin{wrapfigure}{r}{4.2cm}
\vspace{-0.5cm}
\begin{mdframed}[backgroundcolor=gray!20] 
{\em ``Earth observations can generate data for monitoring a number of SDG targets and indicators. Deep Learning contributes extracting meaningful and consistent information.''}
\end{mdframed}
\vspace{-0.5cm}
\end{wrapfigure}
Departing from research laboratories, EO and DL have nowadays the opportunity to contribute to some of the most pressing global societal challenges, such as those identified by the 2030 Agenda for Sustainable Development \new{}{\cite{UnitedNationsGeneralAssembly2015TransformingDevelopment}}. The United Nations (UN) has defined a set of seventeen sustainable development goals (SDGs) as a plan of action to reach peace and prosperity for all people on our planet by 2030. The goals are related to social, economic, and environmental challenges and provide a blueprint for shared action. It is recognised that eradicating poverty in all its forms and dimensions is the greatest global challenge and an indispensable requirement for sustainable development. Each of the seventeen goals has a set of targets and indicators to measure, monitor, and report the progress of each country. The global framework established by the UN is designed around 169 targets and 232 indicators, representing the first truly data-driven framework in which countries can engage with evidence-based decision making and policy development \cite{EuropeanSpaceAgencyESA2018SatelliteGoals}. The 2030 agenda recognizes that \textit{if you can’t measure it, you can't manage it}, thus emphasizing the importance of objective, accurate and trustworthy information for decision making. This approach requires using multiple types of data such as traditional national accounts, household surveys and routine administrative data, as well as new sources such as EO data for the extraction of updated geospatial information. 

The role of EO in support of the SDGs has been recognized and facilitated by international organizations such as the group on Earth observation (GEO), the committee on Earth observation satellites (CEOS), and the European space agency (ESA) \cite{EuropeanSpaceAgencyESA2018SatelliteGoals, GEO2017EarthDevelopment, EuropeanSpaceAgencyESA2020EarthIndicators}. EO can provide continuous temporal information over the globe, capturing the sustainability of the developments underpinning the SDG framework. Satellite, airborne, and unmanned aerial vehicle (UAV) acquisitions provide data at multiple scales for monitoring the state of natural ecosystems, natural resources, oceans, coasts, land, built infrastructure and their change over time. EO data are spatially and temporally consistent, allowing for effective comparison of the results among different countries and in different years. EO data are also complementary with traditional statistical methods, offering a source of information to cross-check the validity of in-situ data measurements (such as survey and inventory data) that are commonly collected by national statistical offices. Moreover, EO can significantly reduce the cost of monitoring SDG targets and indicators with respect to traditional data collection methods. 
According to the ``Compendium of EO contributions to the SDG targets and indicators'' recently released by ESA \cite{EuropeanSpaceAgencyESA2020EarthIndicators}, 34 SDG indicators can be either directly (17 indicators) or indirectly (17 indicators) informed with space-based EO data across 29 targets and 11 goals. Table \ref{tab:SDG-table} summarizes where EO data can contribute to SDG targets and indicators, providing examples of EO applications in support of monitoring the progress and achieving the goals. 
The ESA analysis \cite{EuropeanSpaceAgencyESA2020EarthIndicators} also recognizes the role of the technical infrastructure for storing and processing big EO data, and in particular the relevance of cloud computing, parallel processing systems, and data cubes. However, the contribution of machine learning and DL towards the SDGs is not equally emphasized for their ability to extract meaningful and consistent information from EO data.

\begin{figure} [t]\begin{center}
 \includegraphics[width=1\linewidth]{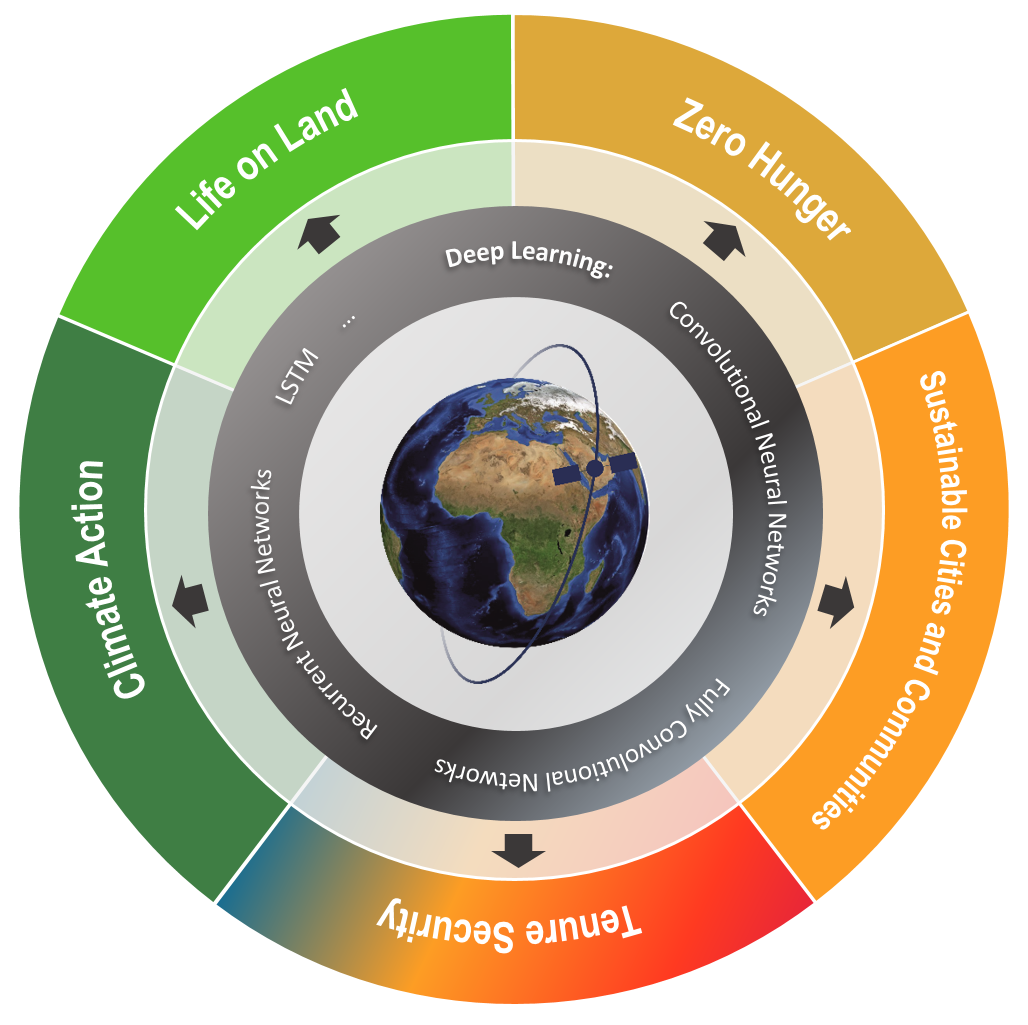}
 \end{center}
 \caption{Framework of the study showing the considered application domains.}
 \label{fig:concept_frame}
\end{figure}

This paper aims to analyze the role and opportunities of DL in EO to support the 2030 agenda for sustainable development 
(Fig. \ref{fig:concept_frame}). Section II provides a review of the recent developments in DL for EO data analysis, placing them in the context of relevant geospatial applications. Section III gives an overview of applications where DL methods applied to EO data contribute towards key SDGs most impacted by the new technological developments. Section IV reflects on the current achievements, challenges, and future perspectives. Section V draws the conclusion.

\begin{table*}[ht!]
\centering
\caption[]{SDG targets and indicators that can be supported by EO-derived products\footnotemark.}
\label{tab:SDG-table}
\begin{tabular}{|c|c|c|p{10cm}|}
\hline
\textbf{SDG}     & \textbf{Targets}    & \textbf{Indicators}     & \textbf{Earth observation application in support of SDG targets and indicators}                                  \\
\hline
\multirow{5}{*}{\includegraphics[width=1.5cm]{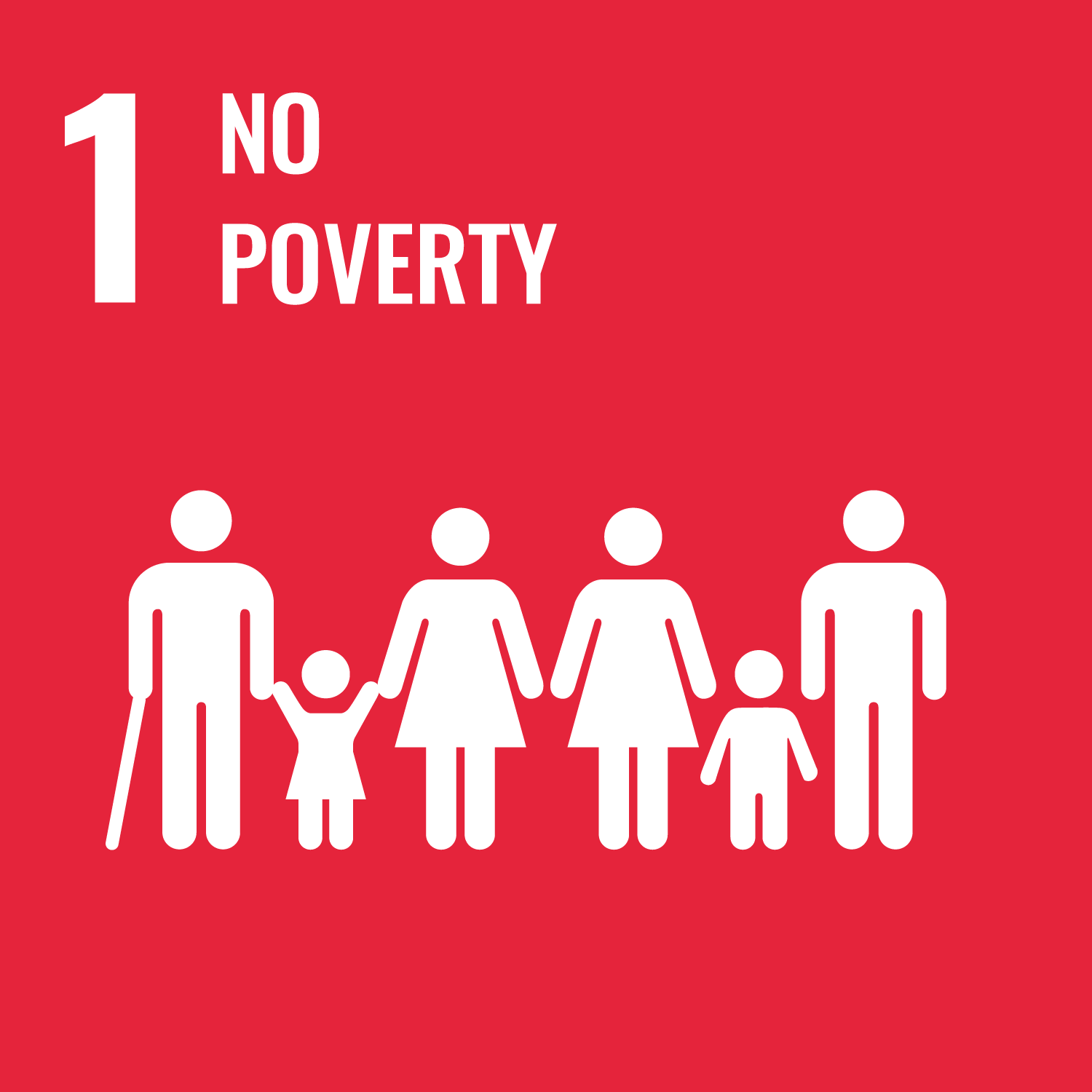}} & 1.4 & 1.4.1, 1.4.2 & Extraction of visible cadastral boundaries and information in support of fit-for-purpose land administration systems \\
         & 1.5 &   & Risk assessment of natural and climate-induced disasters, early warning and post-event damage assessment     \\ 
         & & &       \\
\hline
\multirow{5}{*}{\includegraphics[width=1.5cm]{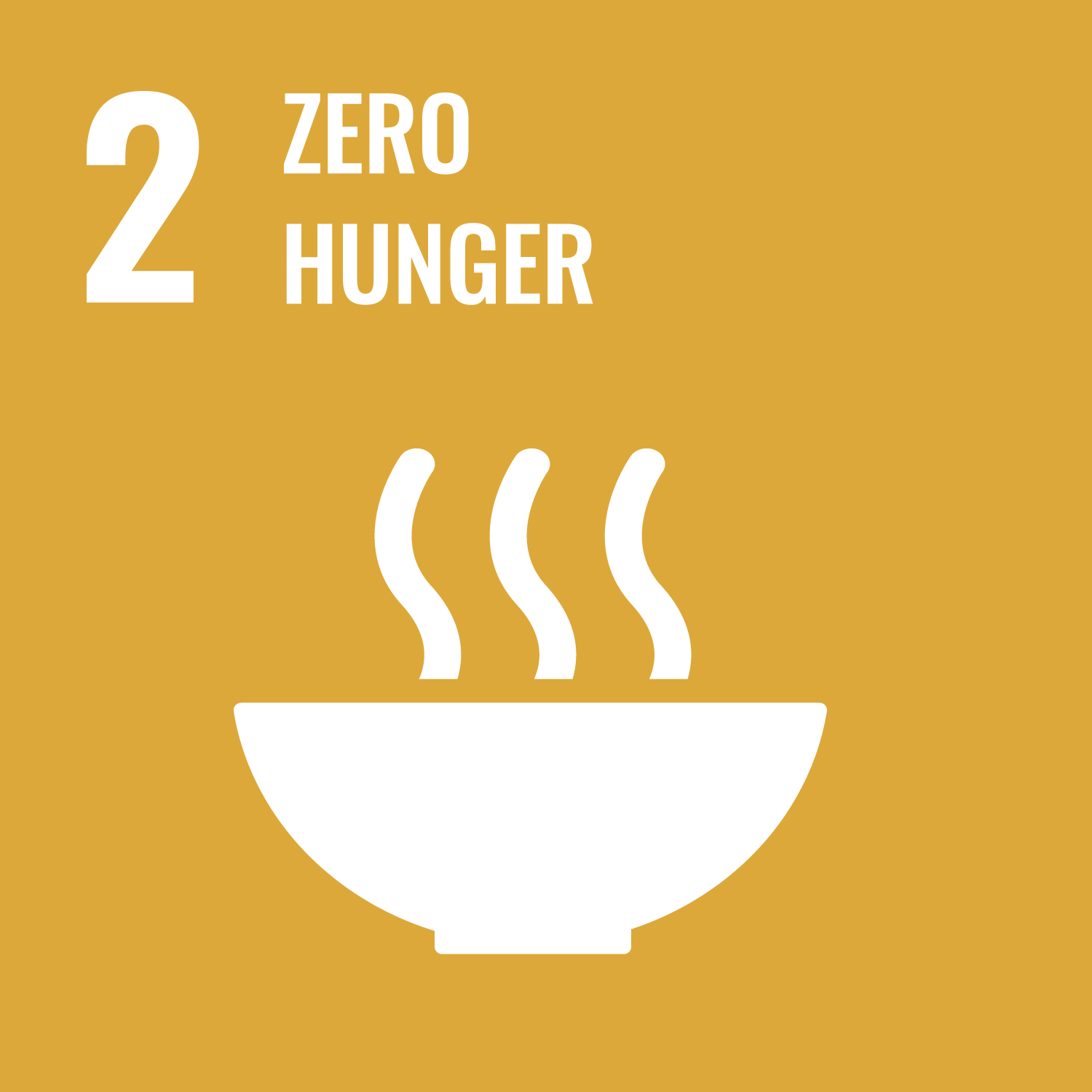}}  & 2.3        & 2.3.1          & Spatial distribution of cropland and smallholder farms; estimation of   agricultural productivity         \\
         & 2.4        & 2.4.1          & Assessment of vulnerability to climate change, extreme weather, drought, flooding and other disasters \\
         & & &                                                   \\
\hline
\multirow{5}{*}{\includegraphics[width=1.5cm]{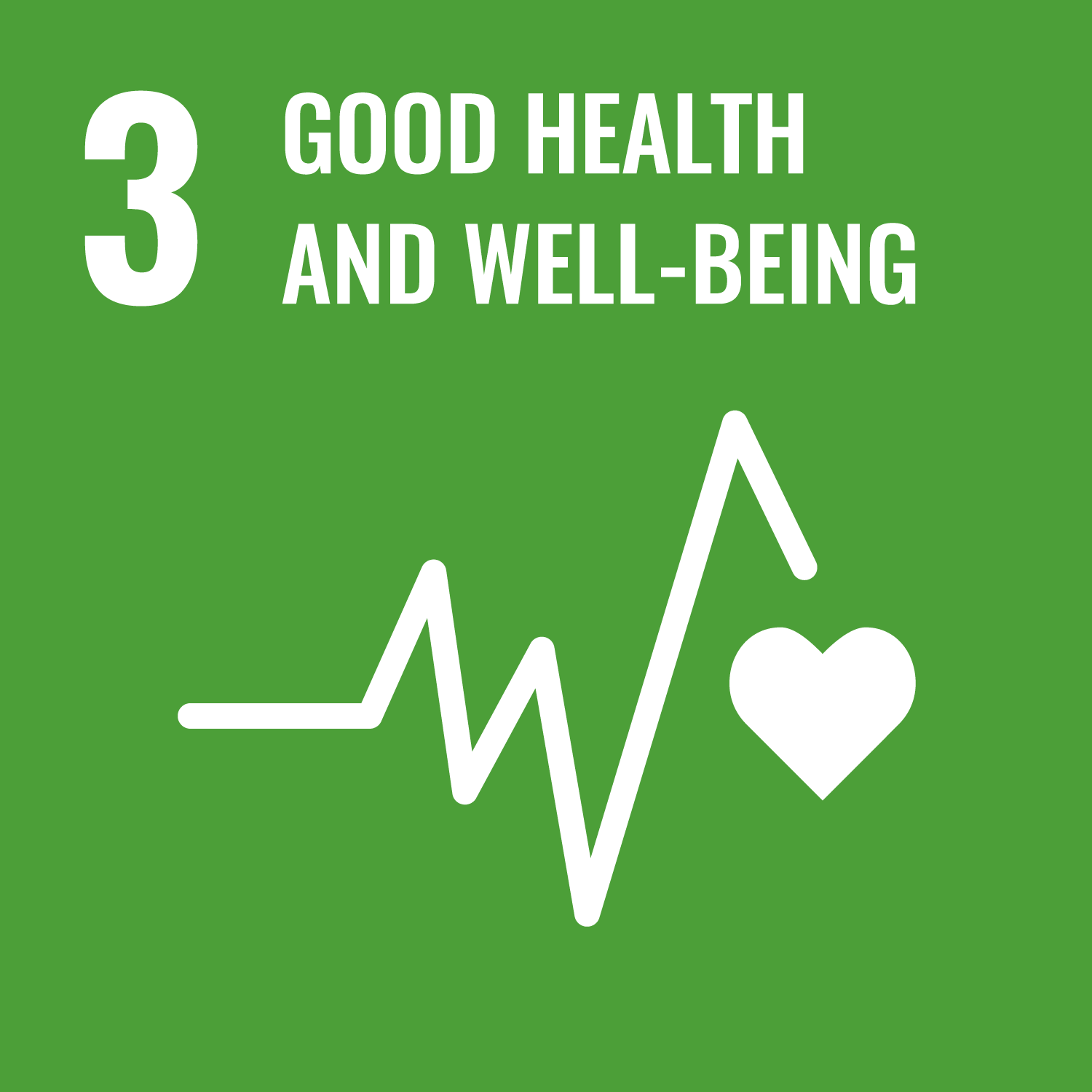}} & 3.3 & & Early warning system for vector-borne disease \\
& 3.6 & & Extraction of road maps and assessment of road conditions (paved/unpaved) \\
& 3.9        & 3.9.1, 3.9.2  & Mapping of hazardous chemicals and pollutants in air, water and soil        \\                                & 3.d & & Geospatial information in support of assessing health risks \\
        & & &           \\
\hline
\multirow{5}{*}{\includegraphics[width=1.5cm]{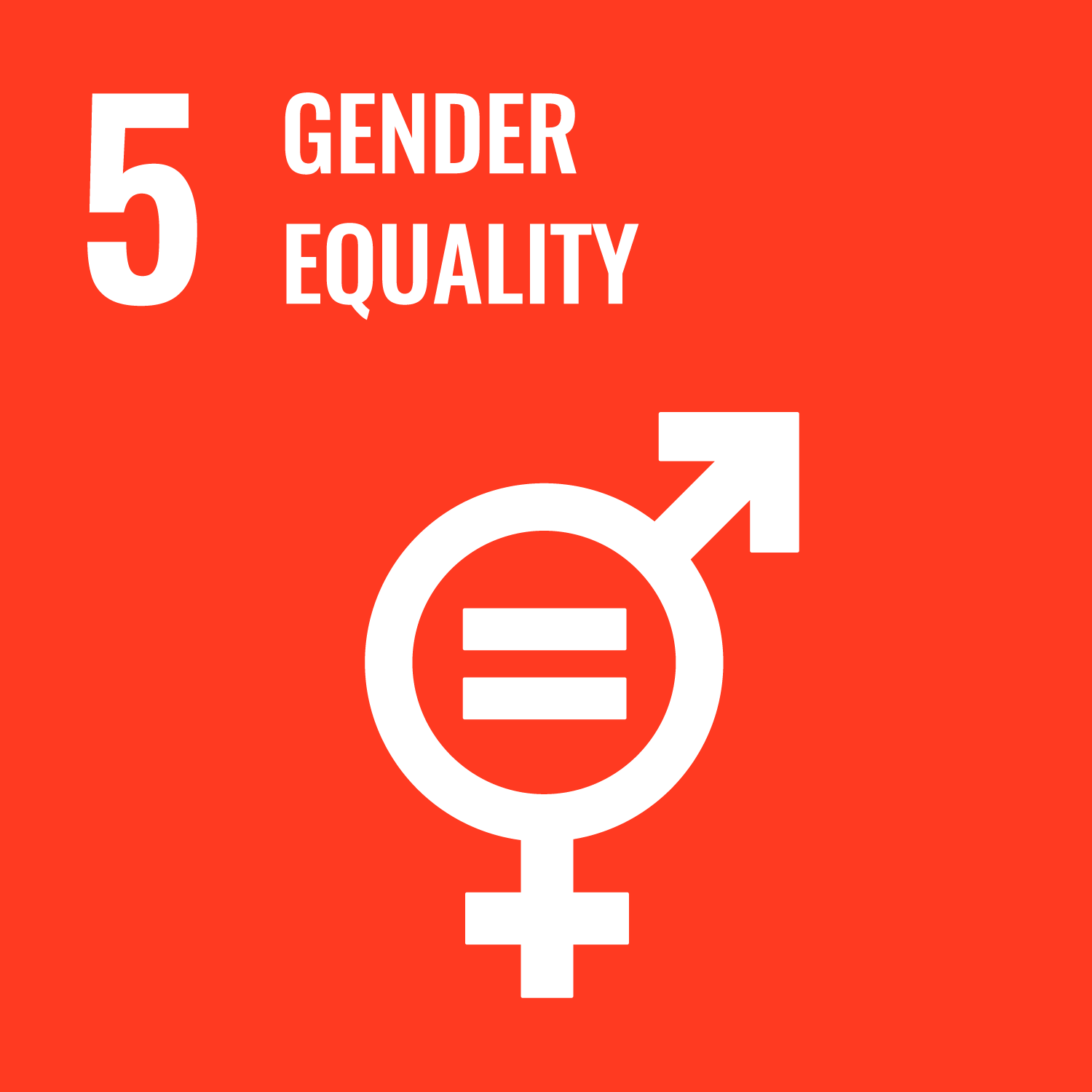}} &
  5.a & 5.a.1 & Extraction of visible cadastral boundaries and geospatial information in support of assessing ownership and secure rights over agricultural land \\
         & & &    \\
         & & &      \\
         & & &      \\
\hline
\multirow{5}{*}{\includegraphics[width=1.5cm]{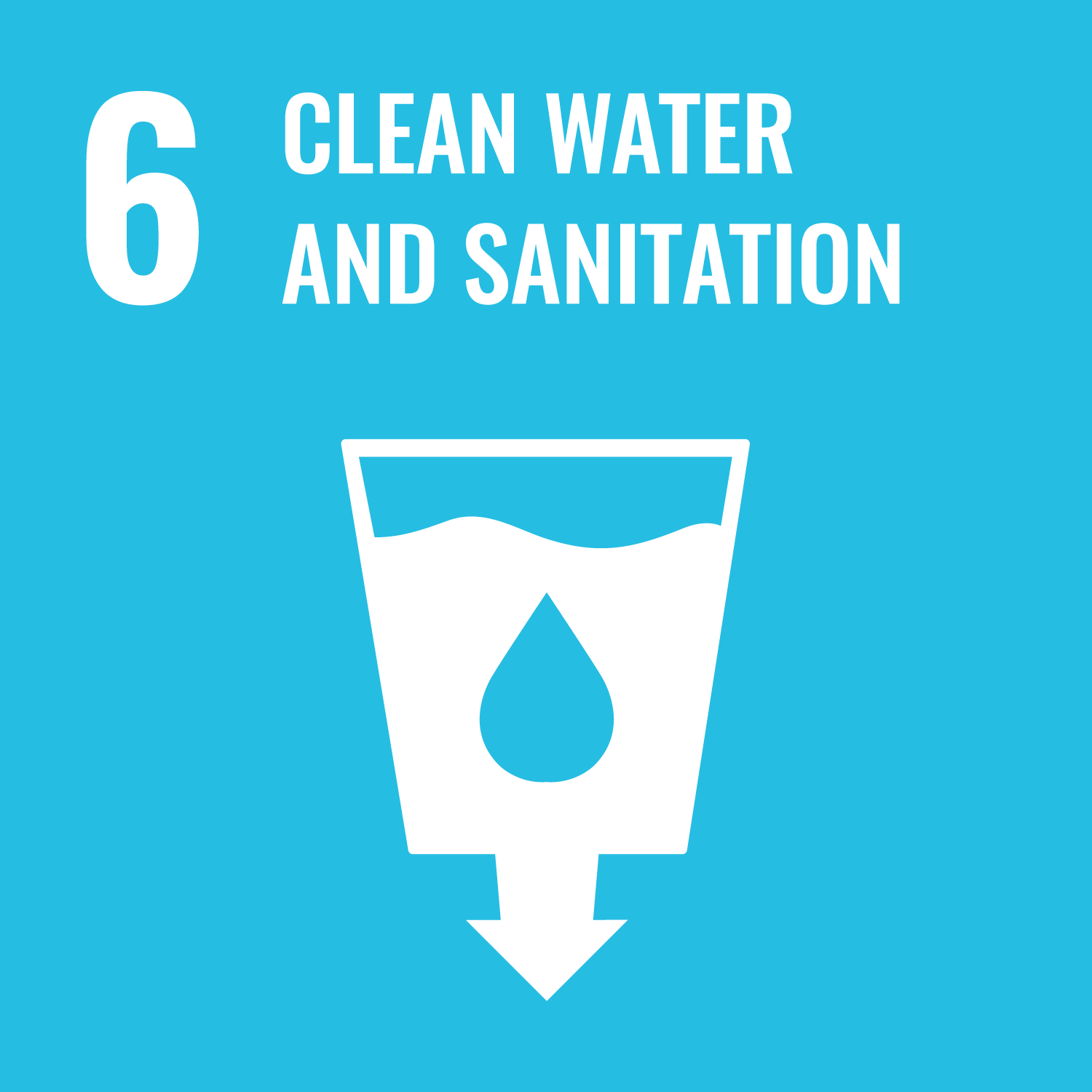}}  & 6.1, 6.3, 6.4 & 6.3.2               & Mapping of water quality and pollutant concentrations                                                                \\
         & 6.5        &                & Geospatial data for runoff modeling; Global rainfall data                                                       \\
         & 6.6        & 6.6.1               & Mapping of water-related ecosystems; Change in the extent of water-related
ecosystems over time                                                                          \\
         & & &      \\
\hline
\multirow{5}{*}{\includegraphics[width=1.5cm]{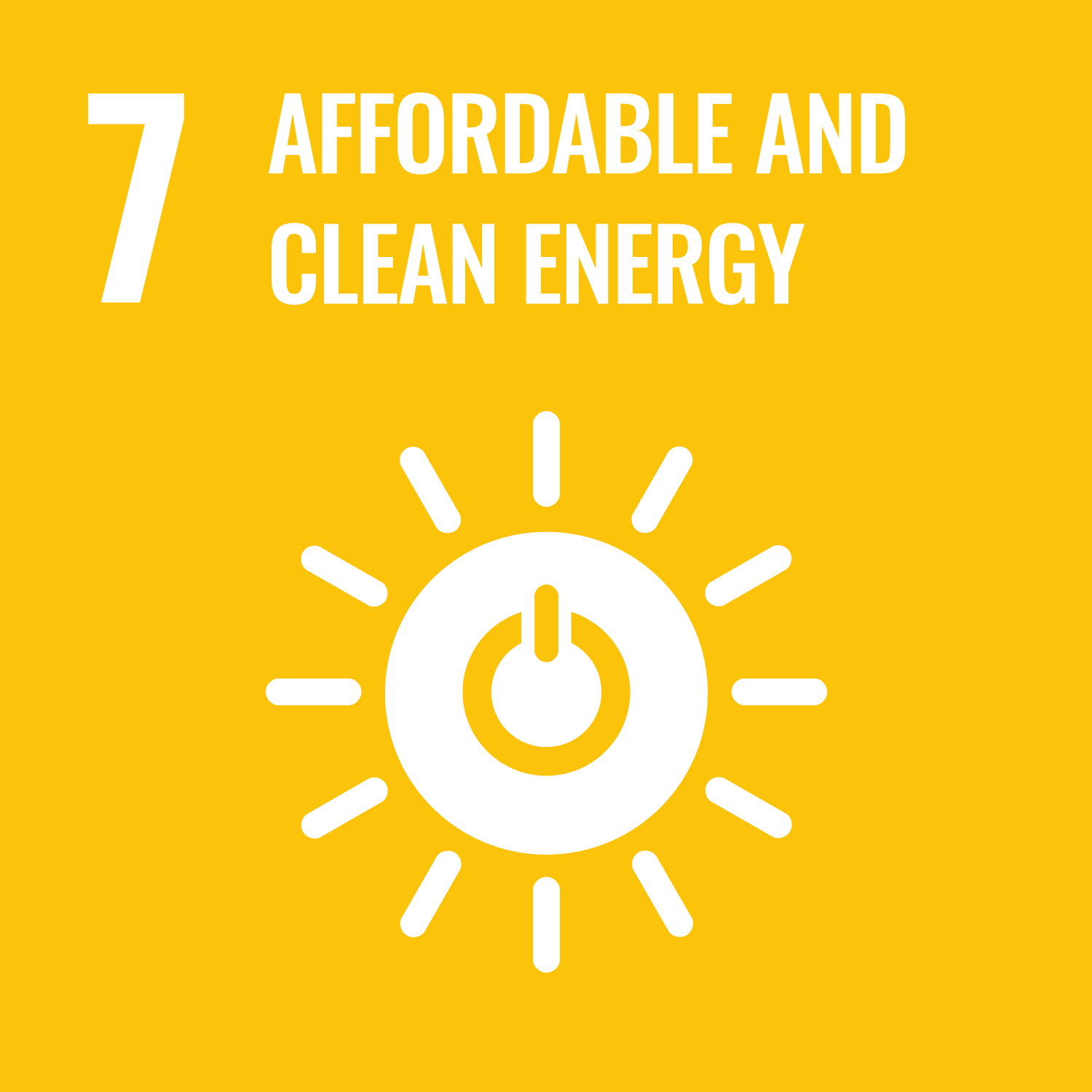}}  & 7.1          &     7.1.1            & Mapping human presence and availability of electricity (e.g., using night-time images)              \\
        
         & 7.2 & & Geospatial information in support of renewable energies      \\
          & & &      \\
         & & &      \\
         & & &      \\
\hline
\multirow{5}{*}{\includegraphics[width=1.5cm]{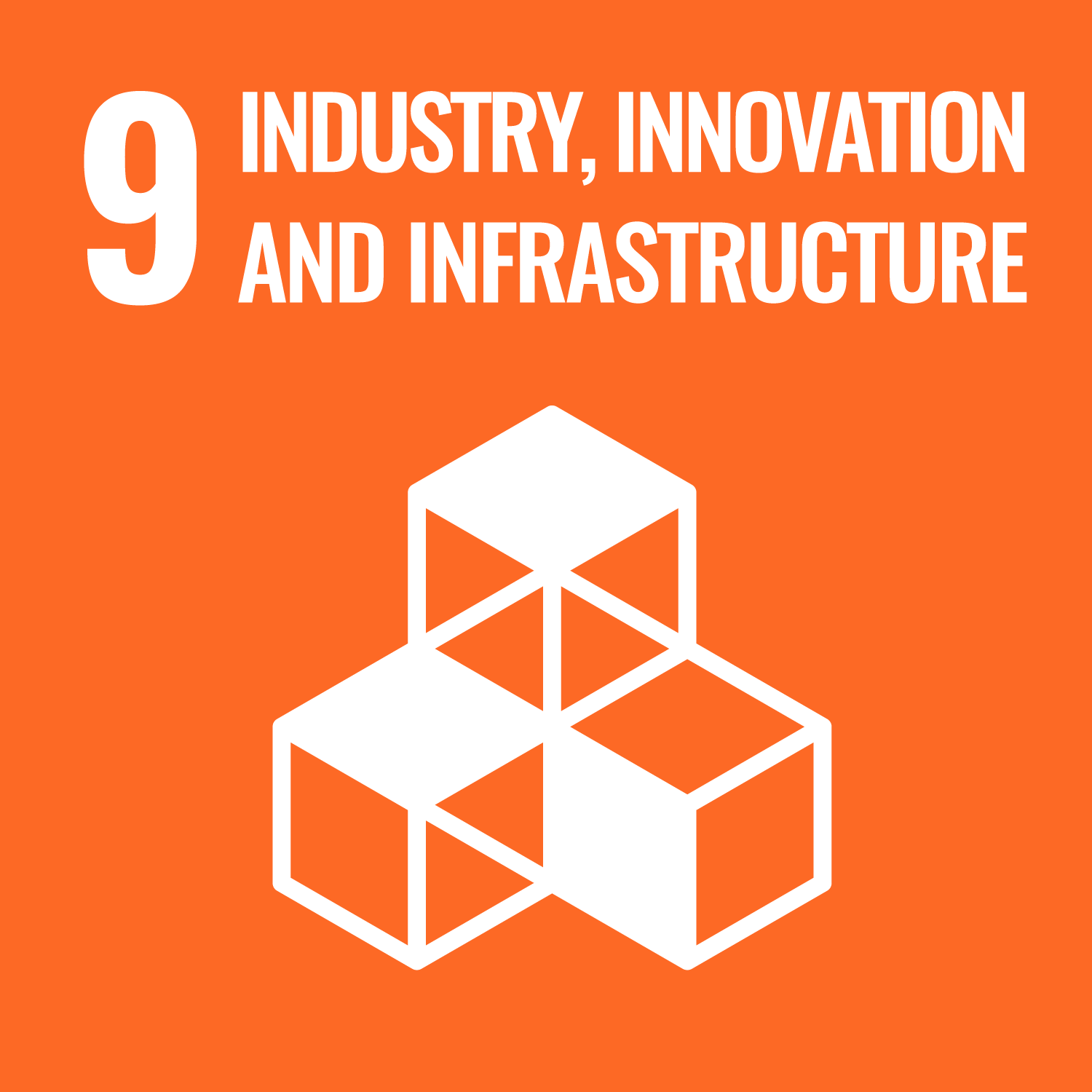}} &
  9.1 &  9.1.1 & Road and transportation network information in rural areas to support assessing accessibility to all-season roads \\
         &            &                &   \\
                 & & &      \\
        & & &      \\
\hline
         
\multirow{5}{*}{\includegraphics[width=1.5cm]{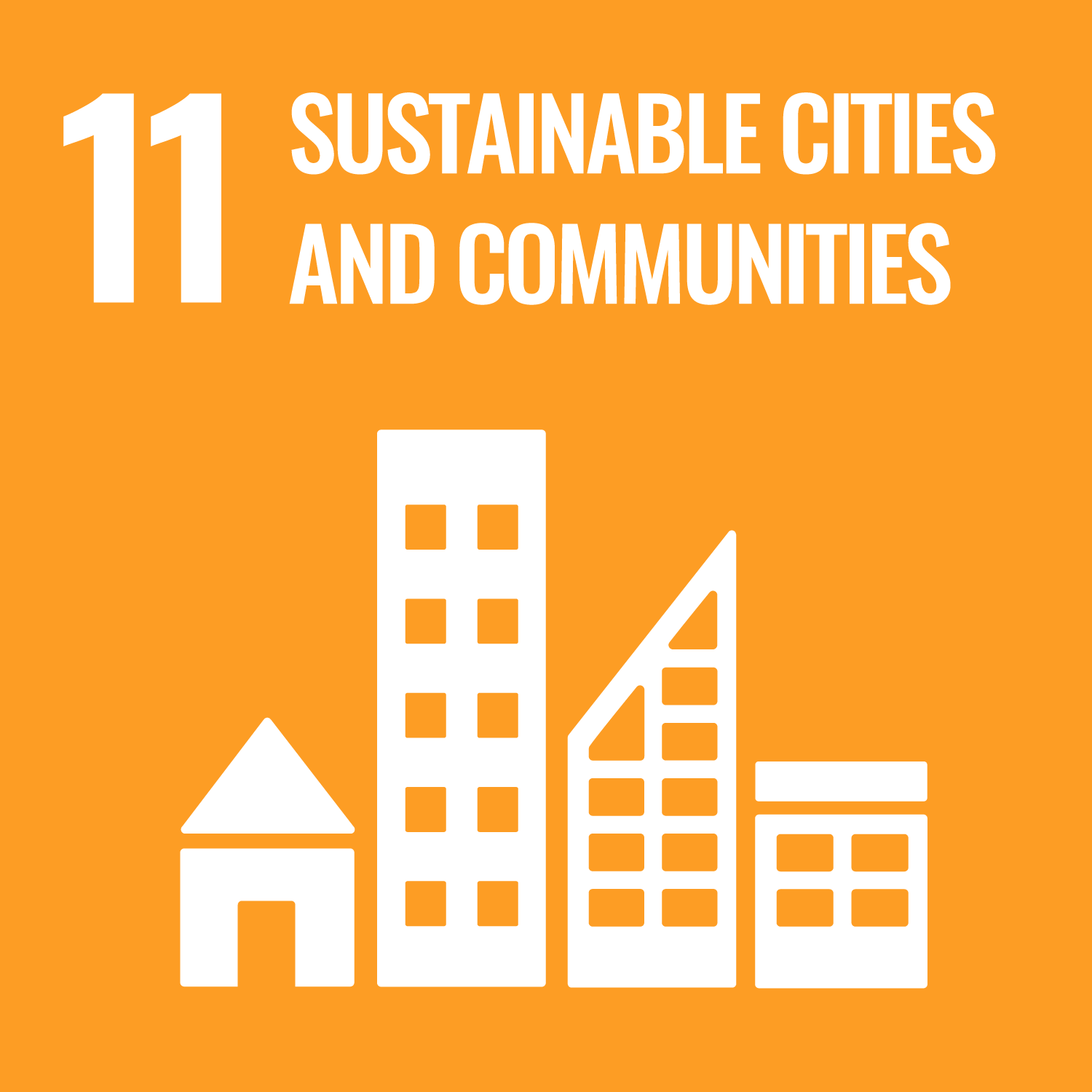}} & 11.1, 11.3 & 11.1.1, 11.3.1 & Mapping of slum distributions and extent, housing quality, density and socio-economic conditions of slum dwellers                                                                   \\
         & 11.2       & 11.2.1         & Road network information for assessing the accessibility to public   transport                            \\
         & 11.4       &                & Geospatial Mapping and monitoring of cultural and natural heritage sites                                    \\
         & 11.5       &                &  Risk assessment and early warning of vulnerable urban areas and disaster-induced damage assessment   \\
         & 11.6       & 11.6.2         & Air quality maps (PM 2.5 and PM 10 concentration); Mapping waste sites                                                                   \\
         & 11.7       & 11.7.1         & Maps of urban green and public open spaces                                                                \\
         & 11.b, 11.c &                & Geospatial information for the development of resilient cities in developing countries          \\
\hline
\multirow{5}{*}{\includegraphics[width=1.5cm]{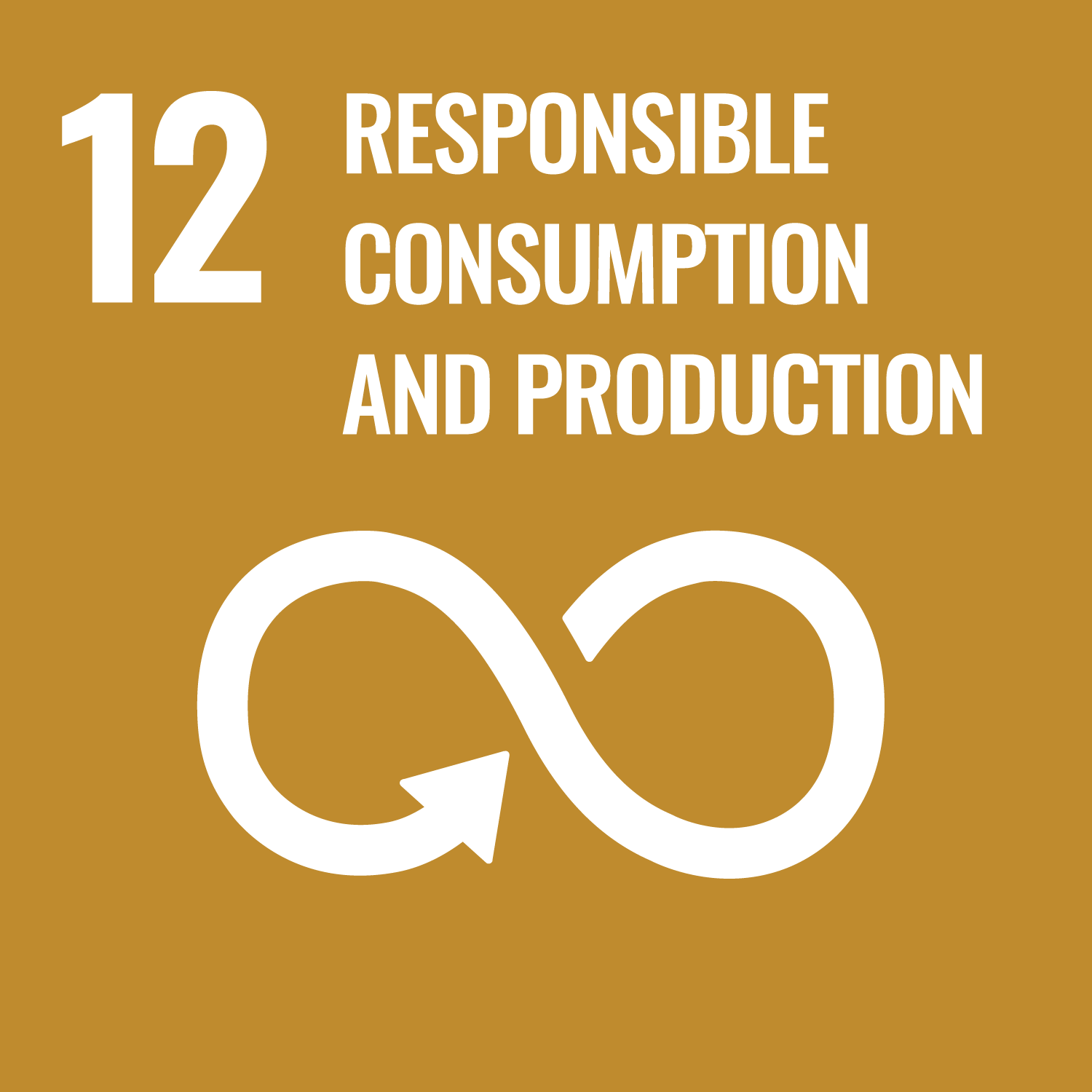}} & 12.2       &                & Maps of natural resources                                                                                 \\
         & 12.4       &                & Information about waste and pollutant released in air, water and   soil                                   \\
         &            &                &                                                                                                           \\
         &            &                &               \\
        & & &      \\
\hline
\multirow{5}{*}{\includegraphics[width=1.5cm]{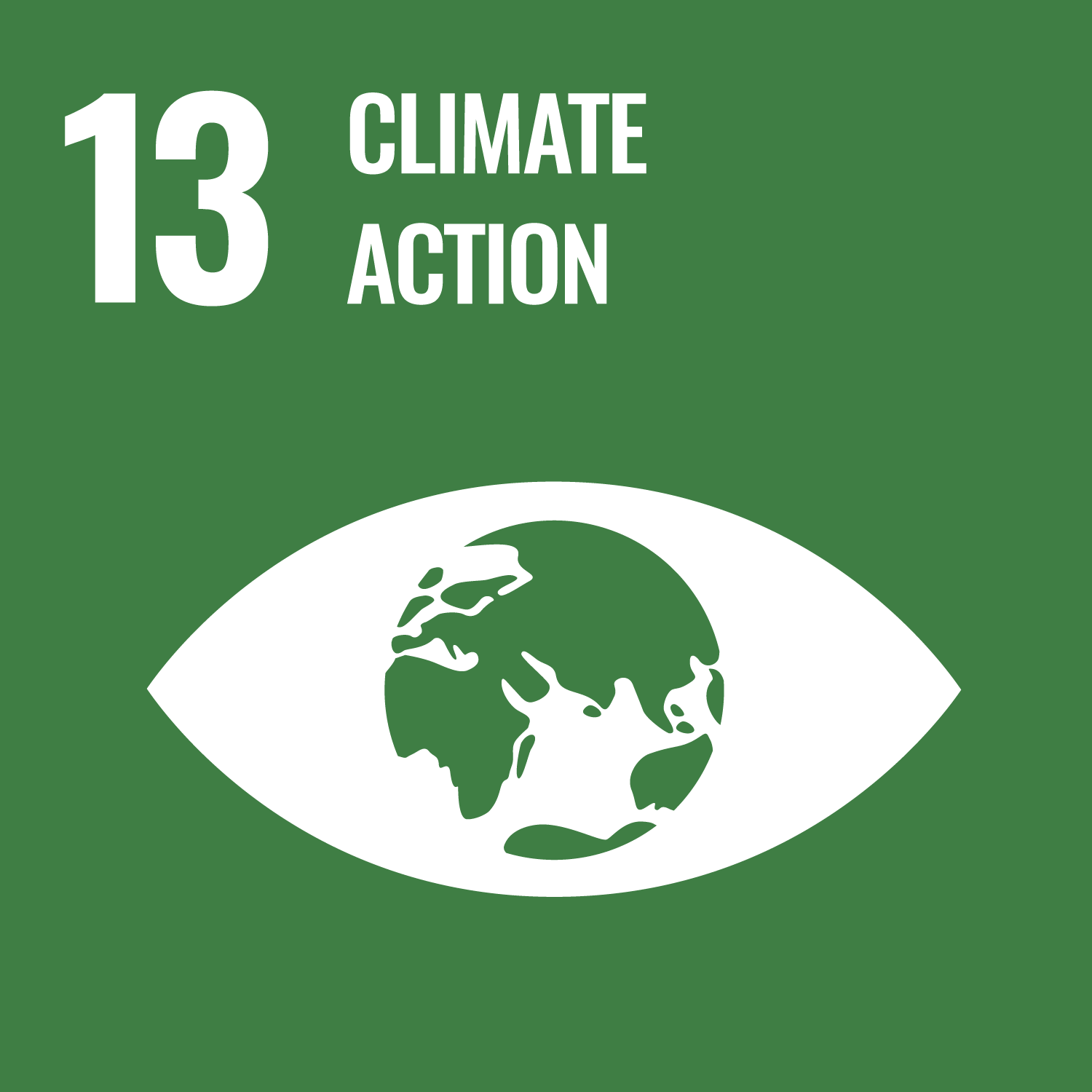}} & 13.1       &                & Risks and damages associated with climate-related hazards and natural disasters                                            \\
         & 13.2       &                & Environmental variables for climate change models                                                                                   \\
         &            &                &       \\
                          & & &      \\
                       & & &      \\
\hline
\multirow{5}{*}{\includegraphics[width=1.5cm]{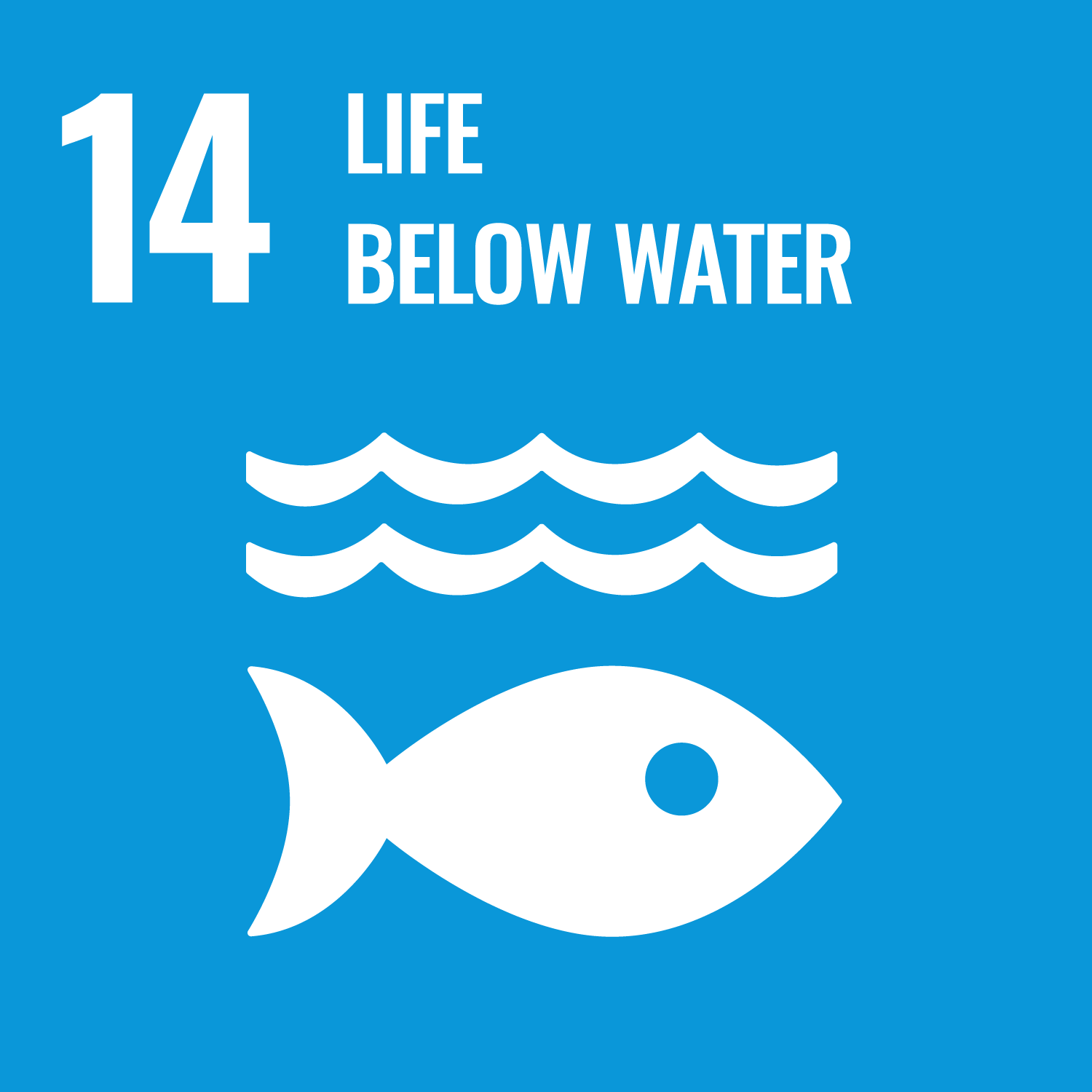}} & 14.1 & 14.1.1         & Coastal eutrophication and floating plastic debris density                                                \\
         & 14.2       &                & Maps of marine and coastal ecosystems                                                                     \\
         & 14.3       & 14.3.1         & Marine acidity (pH)                                                                                       \\
 &
  14.4 &
  14.4.1 &
  Geochemical (chlorophyll concentration) and geophysical analysis (sea surface temperature and ocean   currents) and forecast for global and regional seas \\
\hline

\multirow{5}{*}{\includegraphics[width=1.5cm]{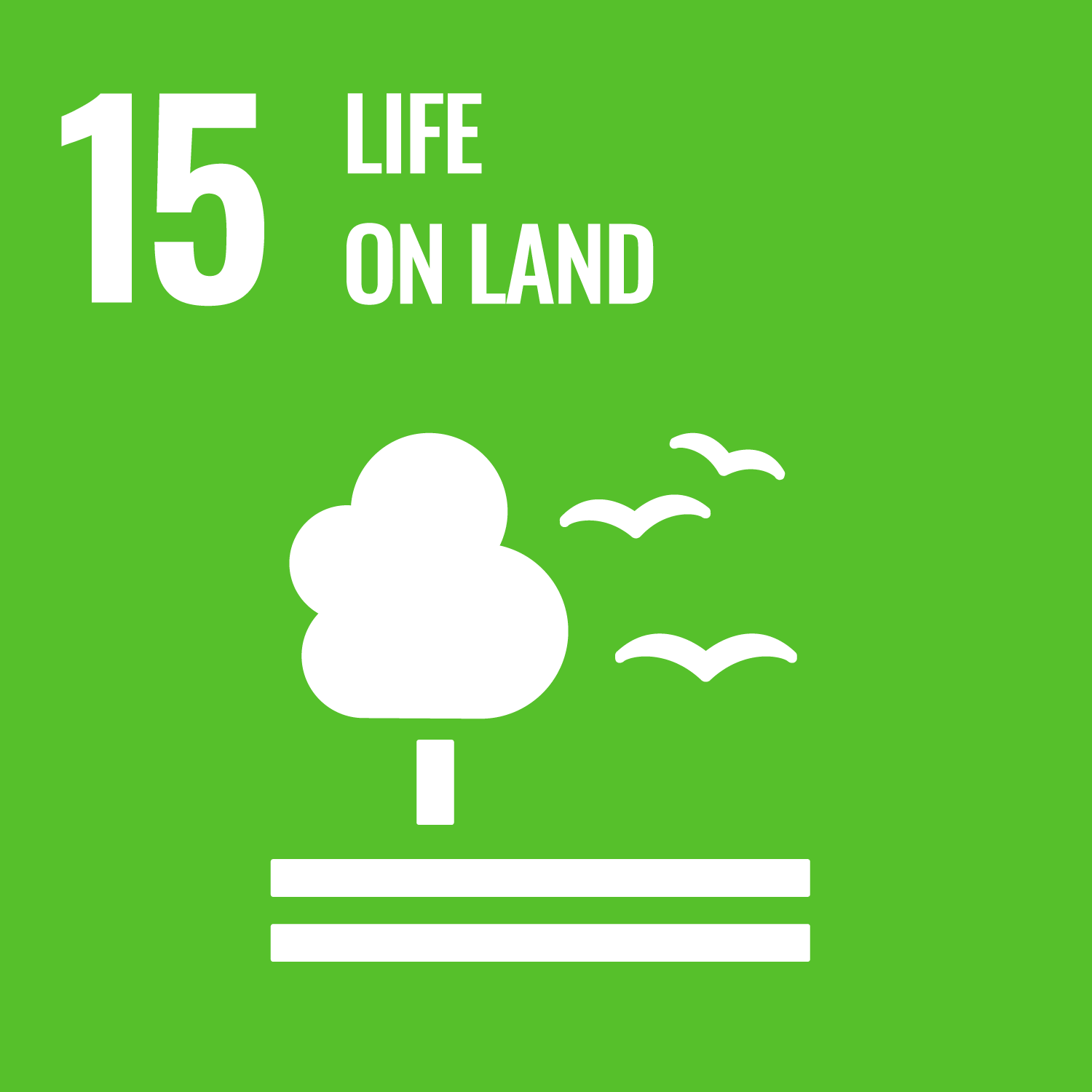}} & 15.1       & 15.1.1         & Forest maps                                                                                               \\
         & 15.2       & 15.2.1         & Forest inventories, deforestation/afforestation maps, wildfire risk assessment                                                      \\
         & 15.3       & 15.3.1         & Maps of deserts and degraded land, prediction of drought and floods                                       \\
         & 15.4       & 15.4.1;15.4.2  & Mountain biodiversity maps                                                                  \\
         & 15.5       &                & Biodiversity maps                                                                              \\
         & 15.7       &                & Wildlife detection to support actions to end poaching and trafficking of protected species\\
\hline
         
\end{tabular}
\end{table*}

\section{Deep learning for Earth observation data}

\footnotetext{Our list differs slightly from those reported in \cite{EuropeanSpaceAgencyESA2018SatelliteGoals, GEO2017EarthDevelopment, EuropeanSpaceAgencyESA2020EarthIndicators}. We focus here on target and indicators that can be more directly supported by EO applications and derived products and is not limited to satellite EO, but also considers other EO platforms.}

Advancements in DL, often based on computer vision research, had a large influence in EO image analysis, resulting in the adoption of DL for a variety of data types and geospatial applications \cite{Zhu2017, Yuan2020DeepChallenges, Ma2019, Hoeser2020ObjectTrends}. 

\subsection{Very high resolution images}
The analysis of very high resolution (VHR) images has been the first to benefit from DL. Given the large amount of spatial information and context contained in VHR images, the extraction of features has always been an active field of investigation~\cite{Fauvel13}. 
With DL, it became possible to learn large dictionaries of convolutional filters directly from data. \new{}{The appearance of publicly available large-scale data sets issued from competitions enabled the appearance of deep networks specific for scene classification \cite{Cheng2018WhenCNNs, Cheng2020RemoteOpportunitiesb} and semantic segmentation ~\cite{rottensteiner2014,DFCA,maggiori2017dataset} of VHR data. The authors in \cite{Cheng2018WhenCNNs} introduced an explicit metric learning regularization term in the loss function to learn more discriminative features.} A wide number of works appeared to process these data sets for classifying land cover at single pixels level: in~\cite{paisitkriangkrai2016semantic}, the authors proposed a hybrid system based on both CNN and traditional descriptor features and then used them in a random forest. In a subsequent paper~\cite{sherrah2016fully}, the author trained two CNN models, one for color and one for height data. In both cases, predictions were provided at the patch level (i.e., a single label is predicted for the whole patch), and a conditional random field was used to smooth the results. 

After these first efforts, papers started to appear applying fully convolutional networks (FCNs) \cite{Long2015FullySegmentation}, providing predictions for each pixel of the patch in one go and greatly reducing the computational cost at inference time~\cite{Aud16,Volpi2017DenseNetworks,Mag17,Mag17c,Persello2017DeepImages}. Since these pioneering works, a large number of papers tackling semantic segmentation have been published in the field and pushed the boundaries of performance on these data sets. Of notable interest are papers that tackled issues such as the integration of prior knowledge
~\cite{Tui17f}, edge information \cite{Marmanis2018ClassificationDetection}, invariances~\cite{Mar17b}, and the explicit inclusion of spatial reasoning \cite{mou2020}. 

\begin{figure} [tbp]\begin{center}
 \includegraphics[width=1\linewidth]{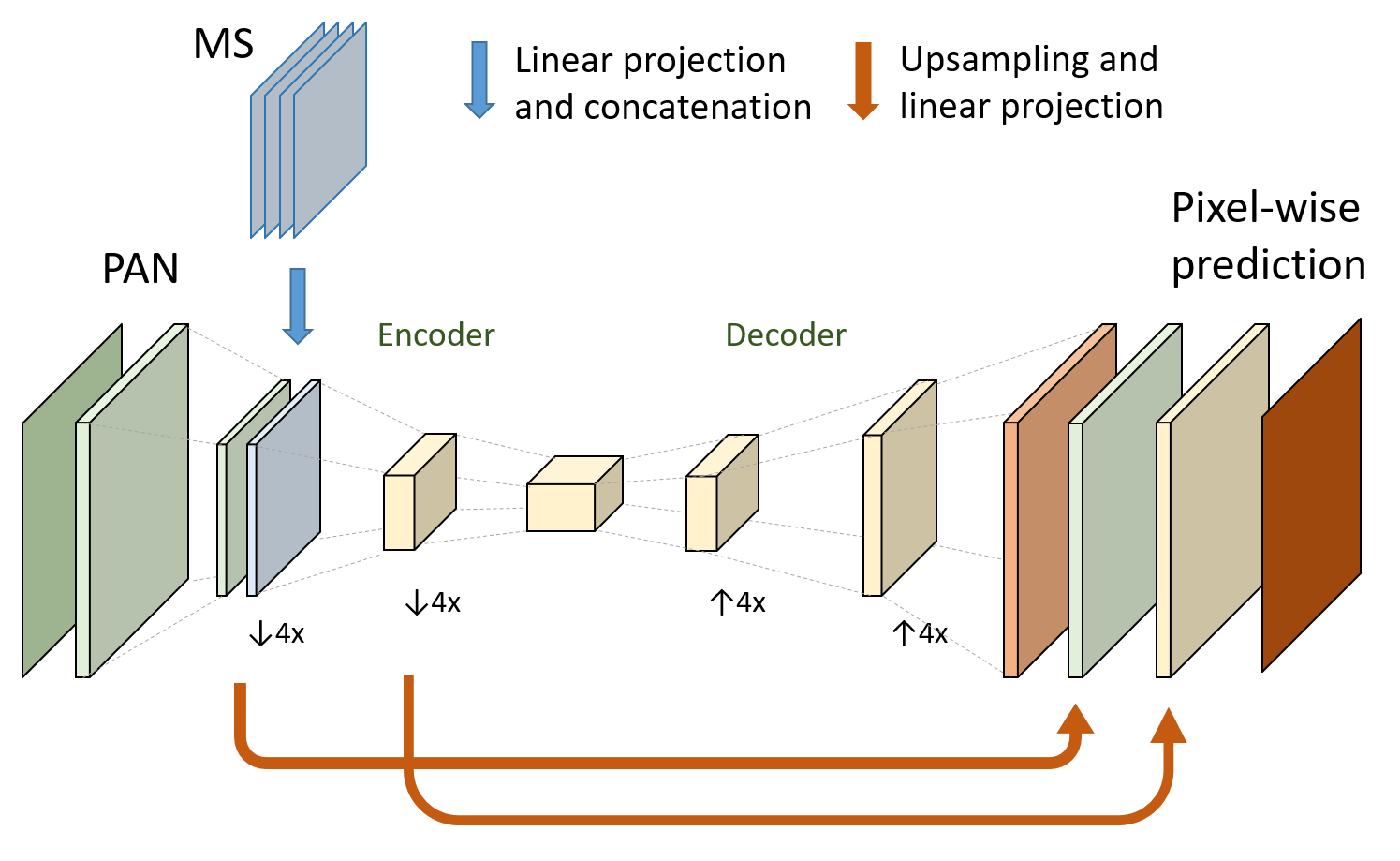}
 \end{center}
 \caption{Architecture of the PAN-MS fusion network (FuseNet) for the pixel-wise classification of VHR multiresolution images  \cite{Bergado2018FuseNet:Classification, Bergado2018RecurrentClassification}.}
 \label{fig:FuseNet}
\end{figure}

Other works looked at methods to fuse the multiresolution bands acquired by most VHR satellite sensors, such as panchromatic (PAN) and multispectral (MS) images \cite{Bergado2018FuseNet:Classification, Shao2018RemoteNetwork, Gaetano2018AImagery}. Bergado et al. \cite{Bergado2018FuseNet:Classification}, introduces a multiresolution FCN, called FuseNet, to perform an end-to-end image fusion and land-cover classification (Fig.~\ref{fig:FuseNet}). This architecture, tailored to VHR satellite data characteristics, resulted in higher performance than methods based on pansharpening. Contextual label information is included in ReuseNet \cite{Bergado2018RecurrentClassification}, a fully-convolutional recurrent network able to learn the contextual label-to-label dependencies that are commonly captured by techniques based on conditional random fields.

Going beyond the human design of CNN architecture, Wang et al. investigate a neural architecture search (NAS) approach \cite{Zoph2017NeuralLearning, Zoph2018LearningRecognition} to automatically design the CNN for the classification of VHR images \cite{Wang2020RSNet:Tasks}.
Unlike other NAS methods based on reinforcement learning or evolutionary algorithms over a discrete and non-differentiable search space, their framework uses a gradient-based method to optimize both architecture and model parameters \cite{Liu2019DARTS:SEARCH}. A switchable module allows for addressing both image classification and semantic segmentation.

\begin{figure} [tb]\begin{center}
\centering
    \begin{tabular}{cc}
    \includegraphics[width=0.35\linewidth]{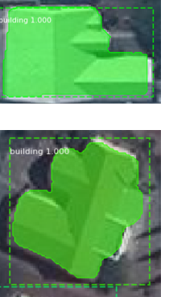} & \includegraphics[width=0.35\linewidth]{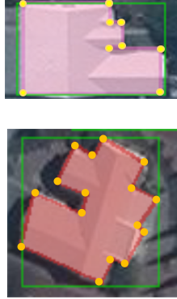} \\
    a) & b) \\
    \end{tabular}
 \end{center}
 \caption{a) Instance segmentation algorithm with raster output (Mask R-CNN) \cite{rhHe2017}. b) Regularized outline extraction with polygonal output \cite{Zhao2021BuildingFramework}.}
 \label{fig:polygon}
\end{figure}

Another promising research line investigates the extension of DL models to the direct prediction of regularized vector outcomes, i.e., outputs that can be immediately ingested in geographic information system (GIS) environments \cite{Mar17c,Li2019TopologicalImages, Girard2020PolygonalLearning, Zhao2021BuildingFramework} (Fig.~ \ref{fig:polygon}). These developments are expected to have many practical uses for building outline delineation, road network extraction, and more in general, for urban planning and monitoring applications in the context of SDG 11.


\subsection{Image time series}
Recurrent Neural Networks (RNN) are a powerful method for modeling sequential data leading to much progress especially in language processing~\cite{sutskever2014sequence, graves2013speech, vinyals2015neural}. Their capability of learning long-range patterns over time make RNNs promising tools for a variety of tasks in remote sensing, too. One important example in the context of SDGs is food security and assessment of famine risk (SDG 2), which calls for large-scale mapping of agricultural activities. RNNs allow learning temporal patterns specific to different kinds of agricultural land use like demonstrated in~\cite{marc_lstm,breizhcrops,pixel_set,star} (see more details in Section \ref{croptypemapping}). 
A cell is the basic building block of an RNN. It combines the data of the current time step in the sequence and the unit's output from the previous time step as two inputs. While RNNs can, in principle, handle sequences of arbitrary and varying length, they are (in their basic form) challenged by long-term dependencies since learning those would require the propagation of gradients over many time steps.
Gated architectures like Long Short-Term Memory (LSTM) cells~\cite{lstm} and Gated Recurrent Units (GRU)~\cite{gruX} aim at mitigating this problem. They use gating mechanisms to store and propagate information over longer time intervals to reduce the vanishing gradient problem.
In general, abstract features are often represented better by deeper architectures~\cite{bengio2009learning}. In the same way that multiple hidden layers can be stacked in traditional feed-forward networks, multiple recurrent cells can also be stacked on top of each other, i.e., the output (or the hidden state) of the lower cell is connected to the input of the next-higher cell, allowing for different dynamics. 

\subsection{Hyperspectral images}   
Hyperspectral images (HSIs) have intensively contributed to SDGs, in particular, SDG 2 \cite{HSISDG2}, SDG 6 \cite{HSISDG6}, SDG 14 \cite{HSISDG14}, and SDG 15 \cite{HSISDG15}.
CNNs might be the most widely used deep architecture for feature extraction and classification due to the utilization of shared weights and local connections, which substantially decrease the number of trainable parameters in such networks compared to its fully connected alternatives. In the literature, 1D- \cite{7882742}, 2D- \cite{zhao2016}, and 3D-CNNs \cite{chen2016} have been employed to extract spectral, spatial, and spectral-spatial information, respectively, from HSI images. 
The high dimensionality of HSIs, which leads to a higher number of trainable parameters compared to gray scale or multispectral images, along with the availability of only a limited number of training samples, make the training stage of such data extremely challenging. To address these issues, some basic strategies such as dropout and weight decay can be used. In addition, four sets of strategies have been investigated to properly train such high-dimensional data with only a limited number of training samples such as dimensionality reduction \cite{ song2018}, data augmentation \cite{Kong2018}, transfer learning \cite{8913592}, and semisupervised or even unsupervised learning \cite{8082108}.


RNNs have also been applied to HSI image analysis. 
By considering the spectral signature of each pixel vector as sequential data, RNN can be applied to a single hyperspectral image for classification \cite{ zhou2019Hyperspectral}. In this context, for each pixel vector, the spectral values are usually fed into RNN from the first band to the last one (this can also be done in a bidirectional way \cite{CRNN}), and the output of the hidden layer at the last band is the extracted spectral feature. In real applications, the sequences' length can be very long (equal to the number of bands), which leads to training difficulties such as gradient vanishing or explosion. To address this issue, a possible solution is to group the spectral bands into shorter sequences \cite{hang2019cascaded} or use LSTM \cite{xu2018spectral} and GRU \cite{7914752}.

\subsection{Synthetic aperture radar}
{Synthetic Aperture Radar (SAR) emits coherent microwave pulses and records the amplitude and phase of their backscattered echo. As an active sensor, it is independent of daylight, and due to the used frequencies, it can penetrate clouds, dust, and to some degree vegetation, soil, ice, and other materials. Applications range from estimating surface characteristics such as roughness and moisture and using polarimetric SAR for land cover/use classification, interferometric SAR for the generation of digital elevation models, and tomographic SAR to estimate height profiles over forests or urban areas. However, SAR has not yet seen the same attention of DL as optical sensors (for a detailed review of DL in SAR we refer the reader to \cite{Zhu2021DeepPerspectives}). The reasons for this are manifold.}
{First, the imaging geometry differs greatly from optical cameras causing effects unknown in optical imagery such as layover or displacement of moving objects. Furthermore, objects' appearance is strongly view dependent (e.g., certain types of backscatter happen only for certain geometric arrangements between sensor and object). Second, SAR records the amplitude and phase of the received backscatter of a coherent pulse and is therefore complex-valued. While the absolute phase of a single-channel SAR image has no direct meaning, the relative phase between two polarimetric channels or two SAR acquisitions is highly important. Since most machine learning methods and frameworks are designed for real-valued data, early approaches to apply DL to SAR data relied on the extraction of real-valued (and mostly hand-crafted) features used as input for the network (e.g. \cite{rhGeng2015}). To address this issue, complex-valued CNNs and FCNs that directly work on the complex-valued data by using complex-valued convolutions, activation functions, as well as loss functions have been introduced in  \cite{rhZhang2017} and \cite{rhMullissa2019}, respectively. A NAS approach is proposed in \cite{rhDong2020} to automate CNN architecture design for SAR data and applied to LC/LU classification. CNNs for scene classification are studied in \cite{rhHuang2020}, and RNN architectures for object detection are investigated in \cite{rhKazemi2019}.}

{The phase of multiple SAR images plays a particular role in interferometric SAR (InSAR) as it relates to changes in height. It is used to generate digital elevation models, monitor earthquakes and volcanoes or general land subsidence. Corresponding networks need to be invariant to constant phase offsets as well as take the cyclic nature of the phase angle into account. CNNs have been used to enhance the quality of measured interferograms \cite{rhHirose2017}, to directly estimate the interferometric phase and coherence \cite{rhSica2020a}, as well as performing phase unwrapping \cite{rhSica2020b}, i.e. the conversion of the cyclic phase into an absolute phase field to estimate topographic heights or deformations. One particular problem in InSAR processing is decorrelation of the two SAR measurements which can be due to several causes including temporal changes as well as volumetric scattering. The latter often indicates vegetation and is used in \cite{rhMazza2019} as input to a U-Net \cite{Ronneberger2015U-net:Segmentation} to derive forests maps.}

{Another effect that has hindered the direct application of methods designed for optical images to SAR data is speckle: a chaotic fluctuation inherent to all measurements based on coherent waves caused by the interference of multiple backscattering in one resolution cell. Speckle reduction has greatly benefited from DL, by using supervised CNN-based denoising \cite{rhWang2017,rhChierchia2017}, a multistream complex-valued FCN \cite{Mullissa2021DespecklingNetwork} or by exploiting approaches such as noise2noise (e.g., in \cite{rhDavis2020,rhMa2020}) that do not require clean data.}

\subsection{Big Geodata fusion}

A sharp increase in the amount of data captured by sensing devices has lead to the \textit{big data} deluge, the creation of the new field of data science and the popularization of DL algorithms to deal with such data \cite{MultiSensor2019}. In a similar manner, the field of remote sensing (RS) has been influenced by an ever-growing number of spaceborne, airborne, and proximate sensing devices such as UAVs to acquire multiscale data from a particular scene. The increase in the number, quality and volume of passive sensing devices has been coupled with the growth in the number of alternative modes of measurement such as airborne light detection and ranging (LiDAR), which generates point clouds representing elevation \cite{Bello_2020}, and SAR sensors \cite{6504845}. Furthermore, the new sources of ancillary data (e.g., data from crowd-sourcing and social media \cite{SALCEDOSANZ2020256}) have been used along with RS devices for a variety of applications in the context of smart cities and smart environment, hazards and disaster identification, or tracking. 

Multi-source data fusion aims to integrate the data of different types, distributions, and sources (can be from a single sensor or from different sensors) by leveraging modality-specific information to improve the performance of the processing approaches compared to a single modality. 
CNN and its variants have significantly contributed to a wide range of multisource data fusion scenarios such as 1) spatio-spectral fusion to produce a fine-spectral fine-spatial resolution image \cite{Masi_2016,8334206}; 2) spatio-temporal fusion to create a fine-spatio-temporal resolution image \cite{8291042,Belgiu_2019}; 3) active (e.g., SAR and LiDAR) and passive (e.g., multispectral and hyperspectral) data fusion mainly to improve classification performance or data matching \cite{HUGHES2020166,8985546}; 4) RS and social media fusion \cite{MultiSensor2019,KANG201844}.

\section{DL Applications contributing to the SDGs}
This section provides an overview of DL and EO applications contributing to the monitoring and achievement of selected SDGs. We focus on applications in the context of 1) zero hunger (SDG 2), 2) sustainable cities and communities (SDG 11), 3) tenure security (multiple SDGs), 4) climate actions (SDG 13), and 5) life on land (SDG 15).  

\subsection{Zero Hunger}

Monitoring agricultural land use and production is essential to achieve zero hunger (SDG 2). It is of high importance for food production, biodiversity, and forestry \cite{Gomez2016}. An increasing world population, climate change, and changes in food consumption habits put yet uncultivated areas under pressure, while leading to intensification in existing agricultural areas \cite{Laurance2014}.
Cropland expansion and intensive use of agricultural areas are often connected with negative ecological impacts like deforestation and biodiversity loss, but also degradation of ecosystem services like ground and surface water quality \cite{Herzog2008, Dise2011}. Therefore, dense, accurate monitoring of agricultural lands plays an essential role for their optimal and sustainable management.

Some of the communities most vulnerable to hunger are smallholder farmers, who dominate agriculture in sub-Saharan Africa with an estimated 51 million farms predominantly characterized by rain-fed production for household consumption \cite{Lowder2016TheWorldwide}. African smallholder farmers often live in poverty in areas prone to natural hazards, where climate change is exacerbating risks of hunger and breakdown of food systems. The large population growth in these areas urgently requires increased production, resilience to natural disasters, improvements in financial services and the governance of food production systems. These improvements are fundamental for defeating hunger and malnutrition, realizing the SDG 2, and in particular target 2.3, which aims to double the agricultural productivity and the incomes of small-scale food producers by 2030. Sustainably increasing the productivity of these agricultural systems and thus improving food security and the livelihoods of smallholder families is a challenge, partly due to a lack of information about these systems.

Knowledge of crop areas and certain land uses is of importance for many political programs that aim to reduce and alleviate the environmental impacts of intensive agriculture too \cite{Gomez2016}. Policy-driven incentives, for instance, foster a particular share of a farm's area to remain extensively used grassland to promote biodiversity, or give subsidies to promote a certain crop mix in the rotation \cite{Finger2012}. Collecting information is traditionally based on farmer self-reporting and spot checking by the authorities in the field, which is laborious, costly, and prone to errors. 
Modern machine learning methods in combination with publicly available satellite imagery provides new possibilities for more accurate spatially dense monitoring of agricultural sites at high temporal resolution and low cost. One particularly promising recent sensor is Sentinel-2, due to its low ground sampling distance (10 m) at a revisit rate of 3-5 days.
In general, the spectral signal of the vegetation as captured by the satellite has specific characteristics as a function of \emph{(i)} soil structure and composition (e.g., soil brightness, soil water content, soil type, etc.), \emph{(ii)} vegetation structure (e.g., canopy cover, Leaf Area Index (LAI), plant height, leaf angle, etc.) and \emph{(iii)} leaf biochemistry (e.g., chlorophyll, water content, nitrogen content, etc.)~\cite{Thenkabail2013}. 
\new{}{Not only does each plant species have its own spectral signature}, but spectral characteristics are also highly dependent on the phenological stage of the plant~\cite{walter2015pheno, anderegg2020spectral}. Instead of merely analysing images at a single point in time, time-series (sequences) of satellite images thus provide significant additional evidence about crop species, and time-series analysis is a standard practice in agricultural RS.

\subsubsection{{Crop type mapping}}\label{croptypemapping}

Crop classification from satellite data has been widely studied in RS. Traditional machine learning approaches with handcrafted features~\cite{inglada2015assessment,vuolo2018much} predominantly rely on vegetation indices like the Normalized Difference Vegetation Index (NDVI)~\cite{foerster2012crop, ustuner2014crop}. Different strategies have been explored to include the temporal evolution as further evidence for classification, such as temporal windows~\cite{conrad2014derivation}, hidden Markov models and dynamic time warping~\cite{siachalou2015hidden,belgiu2018sentinel}, and conditional random fields~\cite{bailly2018crop}. These traditional machine learning models have in common that they struggle to represent the complex spatio-temporal dynamics of spectral features. 
\begin{wrapfigure}{r}{4.3cm}
\vspace{-0.5cm}
\begin{mdframed}[backgroundcolor=gray!20] 
{\em ``Machine learning in combination with open EO data provides new possibilities for monitoring agricultural sites at low cost''}
\end{mdframed}
\vspace{-0.5cm}
\end{wrapfigure}
\begin{figure*}[t!]
\centerline{
\includegraphics[width=8cm]{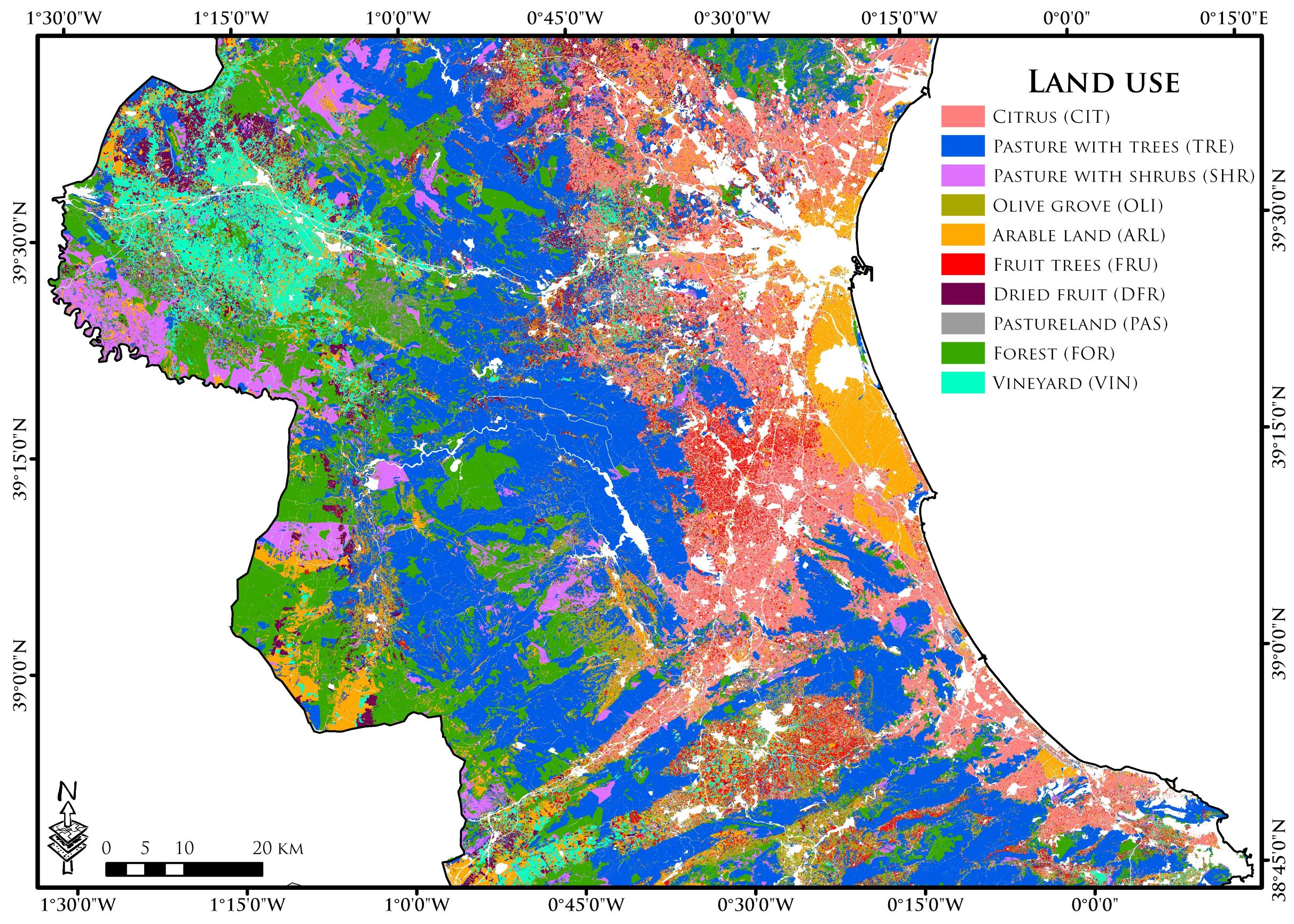}
~~~~~~\includegraphics[width=8cm]{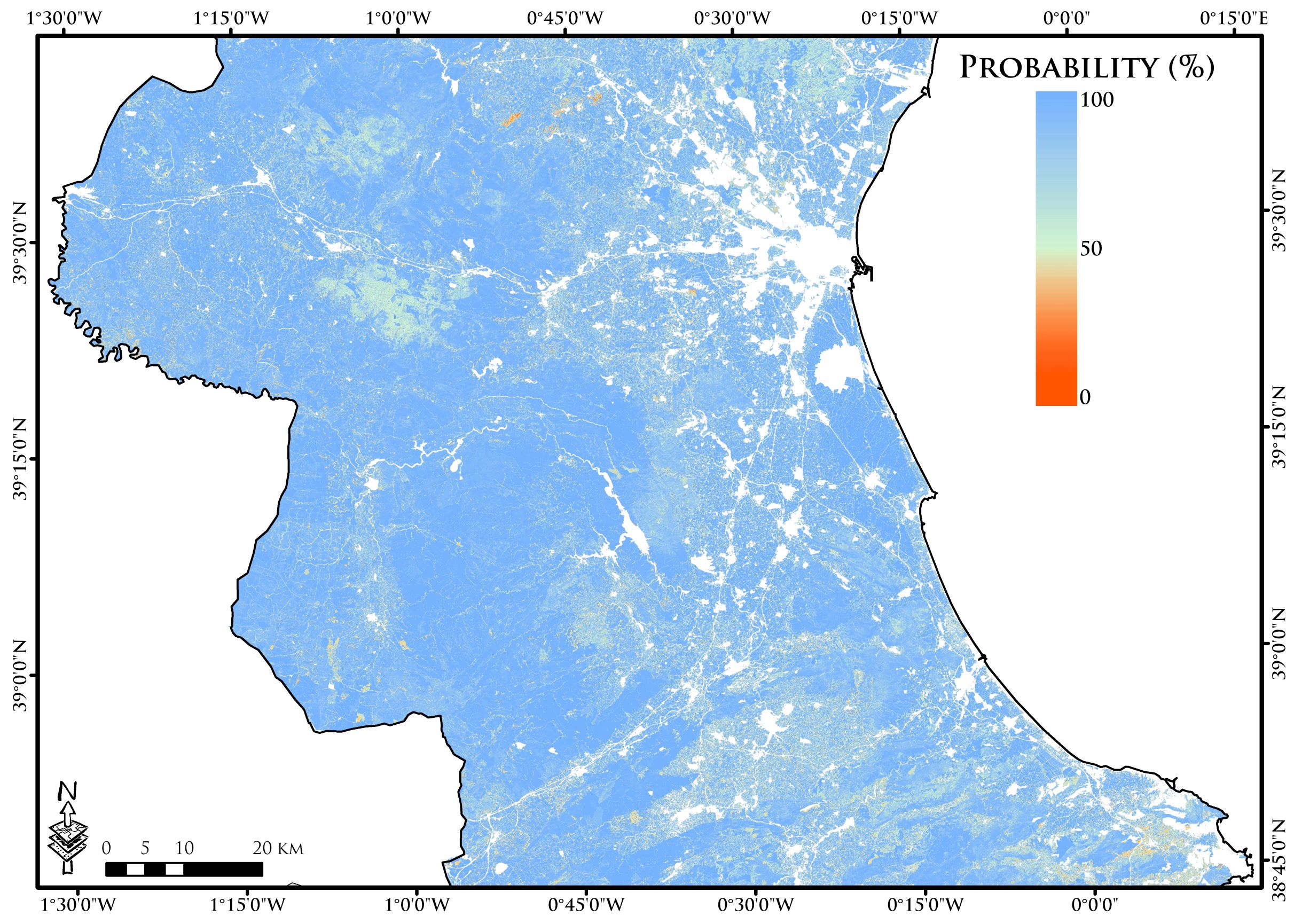}}
\caption{Example of spatio-temporal deep learning for crop monitoring from high resolution images. A bi-LSTM deep network model was used to exploit the spatial regularities and the temporal dimension of a sequence of Sentinel-2 time series to predict the land use over a number of classes of interest (left) and the pixel probability map (right). Figure adapted from \cite{campos2020understanding}. }
\label{fig:cropmap}
\end{figure*}
Recent DL models no longer rely on hand-engineered features to encode spectral, spatial, and temporal patterns. They can learn very complex, highly non-linear relationships if given sufficient labeled training data and computational resources. 
The authors in \cite{rodriguez2018} propose to use a CNN that combines cross-entropy and regression losses for simultaneously mapping and counting oil palm, coconut palm, and olive trees at a country-scale. In \cite{marc_lstm}, the authors use an RNN with LSTM to encode temporal dependencies in the data, while in \cite{bigru} the results on the same data set are improved by encoding both temporal and spatial dependencies via convolutional LSTM. In \cite{fcn}, satellite images are first processed individually with a CNN to obtain per-image features; then temporal dependencies between these features are modeled with a separate RNN. Further options are temporal CNNs that combine features also across time with convolutions~\cite{tempCONV}, or models that use the attention principle~\cite{transformer} to aggregate information across time~\cite{breizhcrops}. The work in \cite{pixel_set} combines pixel-set encoder and transformer~\cite{transformer} and shows improved performance over RNN-based approaches. The authors in~\cite{star} build a deep RNN with a new cell structure termed STAR that trains better than LSTM- and GRU-type models and is more parameter-efficient. This makes it possible to train deeper models, which translates to improved performance across a range of sequence modeling tasks. A recent alternative to RNN approaches for crop mapping are neural ordinary differential equations (NODE) that can interpolate in case of missing data~\cite{metzger2020crop} due cloud coverage, for example. 
Finally, recent approaches have considered spatio-temporal bi-LSTM architectures to fully exploit information of long time series of high-resolution Sentinel-2 data to classify different crop types (rice, fallow, barley, oat, wheat, sunflower, and triticale) \cite{campos2020understanding}, see Fig.~\ref{fig:cropmap}.

\subsubsection{{Delineation of field boundaries}}
\begin{figure*} [tbp]\begin{center}
 \includegraphics[width=1\linewidth]{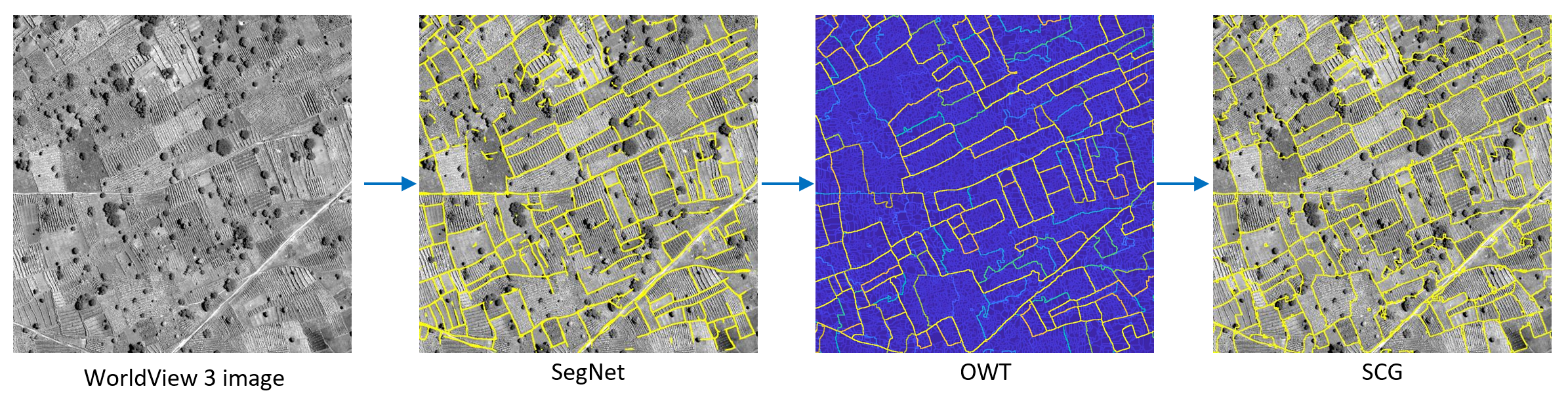}
 \end{center}
 \caption{DL workflow for field boundary delineation in smallholder farms from VHR imagery \cite{Persello2019DelineationGrouping}. SegNet is first applied to extract (fragmented) contours of the agricultural fields. The oriented watershed transform (OWT) is then utilized to extract a hierachical segmentation. Finally, the single-scale combinatorial grouping (SCG) algorithm globalises local cues using an efficient implementation of normalized cuts and explores the combinatorial space of the segmentation hierarchy to generate regions that are likely to represent complete fields.
 }
 \label{fig:field_boundary}
\end{figure*}

Field boundaries are essential for digital agricultural services enabling the estimation of cropland areas, to aggregate and record specific information in a spatial database such as crop grown, soil type, yield, application of pesticide and fertilizer. Moreover, they facilitate the extraction of land tenure boundaries for recording land rights in cadastral systems (see Section \ref{tenure security}). 
Early research on field boundary delineation from EO data has focused on unsupervised techniques based on edge detection or segmentation \cite{Rydberg2001IntegratedImages, Yan2014AutomatedData, Graesser2017DetectionImagery}. These approaches are typically applied to areas characterized by intensive agriculture with large plots using medium resolution images.
However, small area fields ($<$2 hectare) represent 40\% of the fields worldwide and make up 70\% of the cropland in Asia and Africa \cite{Lesiv2019EstimatingCrowdsourcing}. The delineation of such fields is extremely challenging since plots are small, irregularly shaped often with indistinct boundaries. In these circumstances, standard techniques fail in achieving the required accuracy. To this end, DL-based strategies have resulted in significantly higher performance \cite{Persello2019DelineationGrouping, Marvaniya2020SmallData}.
An approach based on SegNet \cite{Badrinarayanan2017SegNet:Segmentation} and combinatorial grouping was proposed in \cite{Persello2019DelineationGrouping} (Fig. \ref{fig:field_boundary}). The FCN is trained to detect field contours discarding irrelevant edges. The detected sparse edges are then used as input to the oriented watershed transform algorithm to extract a hierarchy of closed segments and iteratively merging adjacent regions based on the strength of their common boundary \cite{Arbelaez2011ContourSegmentation}. The final segmentation is obtained by applying the single-scale combinatorial grouping algorithm that explores the segmentation hierarchy to generate accurate field segments \cite{Pont-Tuset2017MultiscaleGeneration}. Promising results are obtained in two study areas in Nigeria and Mali. Marvaniya et al. present a multi-stage approach that uses a combination of DL for edge detection and a sequence of post-processing steps for improving the results \cite{Marvaniya2020SmallData}.

Other recent DL-based solutions include \cite{Garcia-Pedrero2019DeepSystem, Waldner2020DeepNetwork, Masoud2019DelineationNetworks, Wagner2020DeepExtraction}. A method based on U-Net and open data from the land parcel identification system of Spain was investigated in \cite{Garcia-Pedrero2019DeepSystem}. 
Waldner and Diakogiannis adopted a multitask approach to tackle the problem. They used a ResUNet-a FCN to identify: 1) the extent of fields; 2) the field boundaries; and 3) the distance to the closest boundary \cite{Waldner2020DeepNetwork}. Using a single monthly composite image from Sentinel-2 as input, their model could accurately map field extent and boundaries. Other notable works have investigated a super-resolution mapping approach \cite{Masoud2019DelineationNetworks} and the combination of neural networks with graph-based growing contours method to extract agricultural field polygons \cite{Wagner2020DeepExtraction}.


\subsection{Sustainable cities and communities}

Cities are the economic hubs of modern nations and home of an estimated 55.3\% of the world’s population. By  2030, urban areas are  projected to house 60\% of the people globally, reaching 68\% by 2050 \cite{Nations2018WorldRevision}. While urbanization creates opportunities for economic developments, it also creates enormous social and environmental challenges. Some of the most pressing issues are the management of natural hazards,  pollution, and the surge of socioeconomic inequalities resulting from excluding the poor from the social fabric.
\begin{wrapfigure}{r}{4.50cm}
\begin{mdframed}[backgroundcolor=gray!20] 
{\em ``Approximately 1 billion people worldwide reside in informal settlements, living in deprived conditions lacking access to essential services''}
\end{mdframed}
\vspace{-0.2cm}
\end{wrapfigure}
According to UN-Habitat, approximately 1 billion people worldwide reside in informal settlements, commonly called slums, living in deprived conditions lacking access to essential services such as safe water, acceptable sanitation, and durable housing \cite{UN-Habitat2015HABITATSETTLEMENTS}.

The rapid urbanization processes in low and middle-income countries contribute to the proliferation of deprived neighborhoods where dwellers live in crowded areas in unhealthy conditions and often without tenure security. In addition to that, these communities are also among the most vulnerable to the effects of climate change and the increasing frequency and intensity of natural disasters such as floods, heatwaves, droughts, landslides, storms, wildfires, and cyclones \cite{UN-Habitat2018AddressingSETTLEMENTS}. 

\subsubsection{Mapping slums and urban poverty}
The 2030 agenda pays particular attention to these global challenges with SDG 11, which aims at inclusive, safe, resilient, and sustainable cities and human settlements. \new{}{The key indicator 11.1.1 requires monitoring} ``the proportion of urban population living in slums, informal settlements or inadequate housing.'' 
Current global statistics show a decline in the percentage of urban population living in slums, but an absolute increase of inhabitants living in such areas \cite{UN-Habitat2015Slum2016}. Nevertheless, official national statistics are often outdated, inconsistent, or simply inaccurate. Small slum pockets are generally neglected, and population counts based on census data subject to large uncertainties, especially in large metropolitan areas \cite{Kuffer2018TheIndicator}. More accurate and globally consistent methods to gather data on the slum population and their socioeconomic conditions are therefore needed. 

\begin{figure} [tb]\begin{center}
\centering
    \begin{tabular}{c}
        \includegraphics[width=\columnwidth]{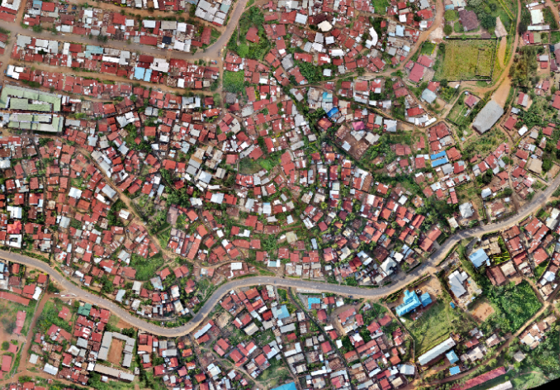}\\
        a) 2D Orthomosaic \\
        \includegraphics[width=\columnwidth]{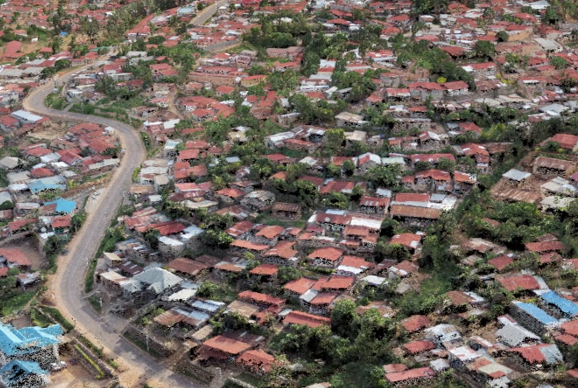}\\
        b) 3D photogrammetric point cloud \\
    \end{tabular}
 \end{center}
 \caption{UAV image acquired over an informal settlement in Kigali, Rwanda. a) 2D Orthomosaic. b) 3D photogrammetric point cloud \cite{Gevaert2017InformalData}.}
 \label{fig:slum_UAV}
\end{figure}

Several studies show the ability of remote-sensing techniques to identify informal settlements, providing a relatively consistent mapping approach applicable over large areas and repeatable in time \cite{Kuffer2016SlumsSensing}. 
Maps derived from VHR satellite data can support SDG 11 and the monitoring indicator 11.1.1 in particular. Detailed 2D and 3D geospatial information extracted from UAV data can support the planning and monitoring of urban upgrading projects, thus contributing to targets 11.b and 11.c \cite{Gevaert2017InformalData, Gevaert2020MonitoringVehicles} (Fig.~\ref{fig:slum_UAV}). Mapping informal settlements can be performed on the basis of physical and morphological characteristics captured by VHR satellite images. Slums are commonly densely built-up areas characterized by small buildings arranged according to irregular layout patterns and lack of green spaces. Extracting these characteristics automatically from images is, however, a difficult task. The spectral information alone is insufficient to discriminate between different urban typologies (formal vs. informal). It is necessary to extract contextual features capable of capturing long-range pixel dependency for distinguishing the different spatial patterns. Conventional machine learning approaches resort to the extraction of texture statistics, local binary patterns, oriented gradients, and segment-based features \cite{Graesser2012ImageLandscape, Kuffer2016ExtractionVariance}. However, these methods depend on several free parameters, which are difficult to optimize and usually set according to user experience. 

The ability of CNNs to automatically learn high-level spatial features results in a streamlined workflow for slum mapping and higher classification accuracy. Mboga et al. apply CNN to detect informal settlements in Dar es Salaam, Tanzania, reporting an accuracy improvement over a support vector machine classifier trained with texture features and local binary patterns \cite{Mboga2017DetectionNetworks}. The authors in \cite{Persello2017DeepImages} introduce FCNs for mapping informal settlements from VHR images. To this end, they adopt an FCN architecture with dilated convolutions (named FCN-DK), therefore capturing long-range pixel dependencies while keeping a limited number of network parameters. The best results are obtained by a network with six convolutional layers using increasing dilation factors. Moreover, they report a significant advantage in terms of computational cost at testing time with respect to patch-based CNN.
Wurm et al. investigate the transferability of an FCN model pre-trained on VHR images to map slums in coarser resolution Sentinel-2 images and SAR data acquired by TerraSAR-X \cite{Wurm2019SemanticNetworks}. They use an FCN-VGG19 architecture adapted from \cite{Long2015FullySegmentation}. Their results show that transfer learning can significantly improve the results on Sentinel-2 while not on TerraSAR-X data.
Wang et al. investigate a U-Net compound model, including dilated convolution operations, to map deprivation pockets in Bangalore, India \cite{Wang2019DeprivationNetworks}. The authors in  \cite{Liu2019TheApproach} use an FCN-based approach to study the temporal dynamics of slums, looking in particular at the temporary slum pockets. The study investigates two change detection approaches based on FCN with dilated convolutions. The first approach uses a post-classification change detection, and the second trains FCNs to directly classify the transition in the land cover classes.

\begin{figure*}[tb]
\centering
\begin{tabular} {ccccc}
& VHR image & Reference map  & Patch-Based CNN & FCN-DK6 \\
&
\includegraphics[width=4cm]{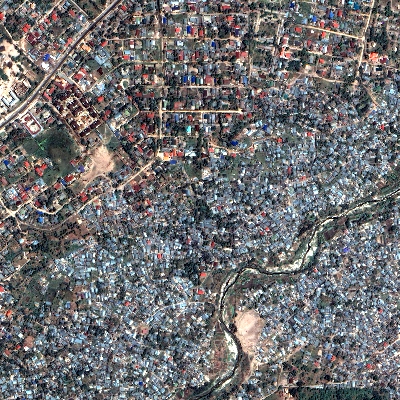}&
\includegraphics[width=4cm]{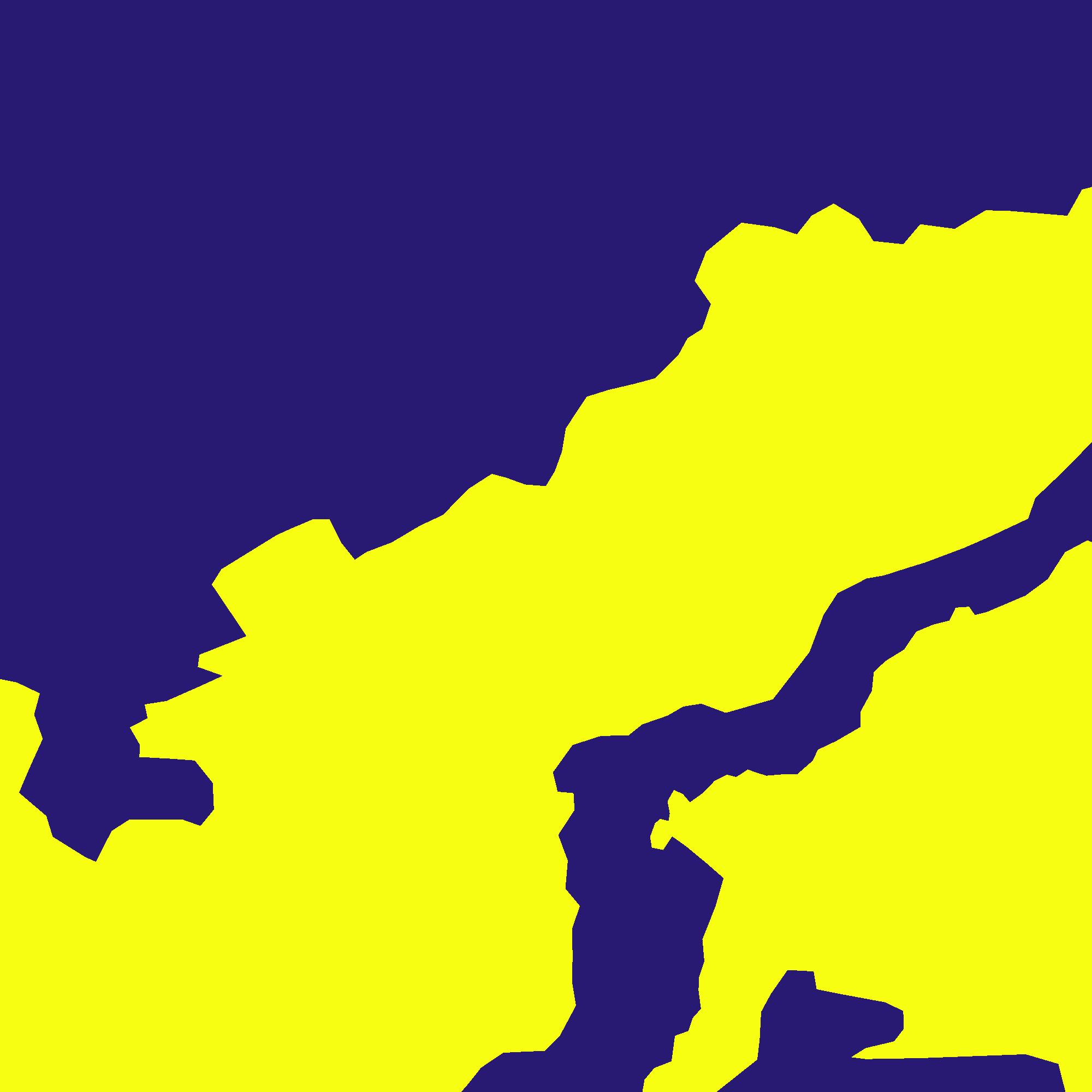}&
\includegraphics[width=4cm]{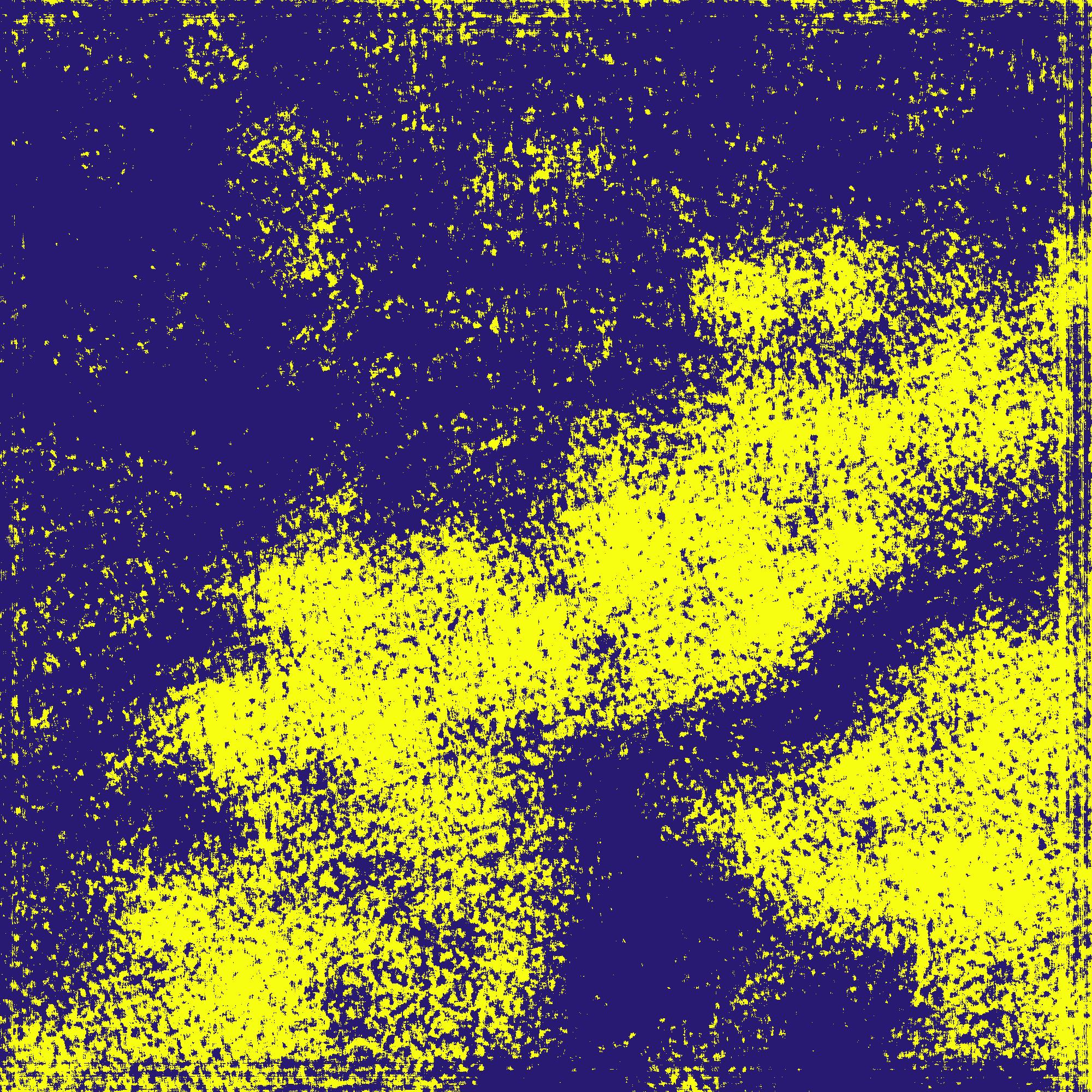}&
\includegraphics[width=4cm]{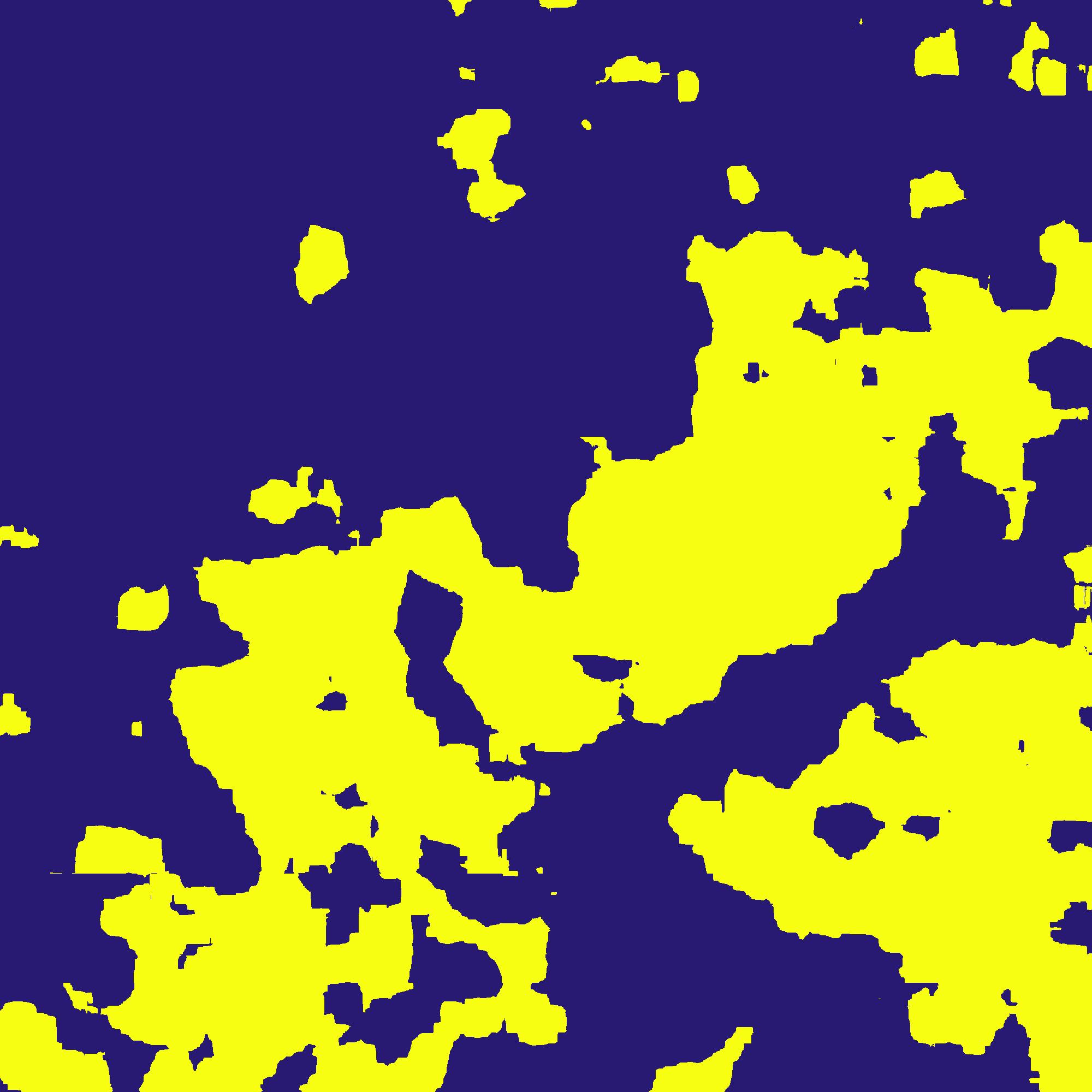}\\
&
\includegraphics[width=4cm]{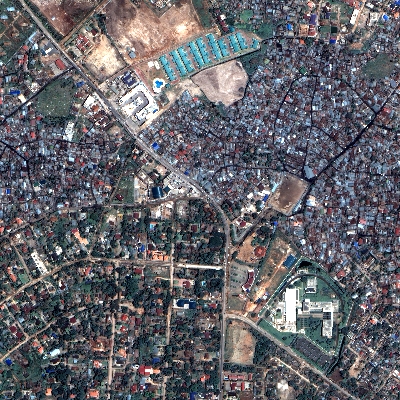}&
\includegraphics[width=4cm]{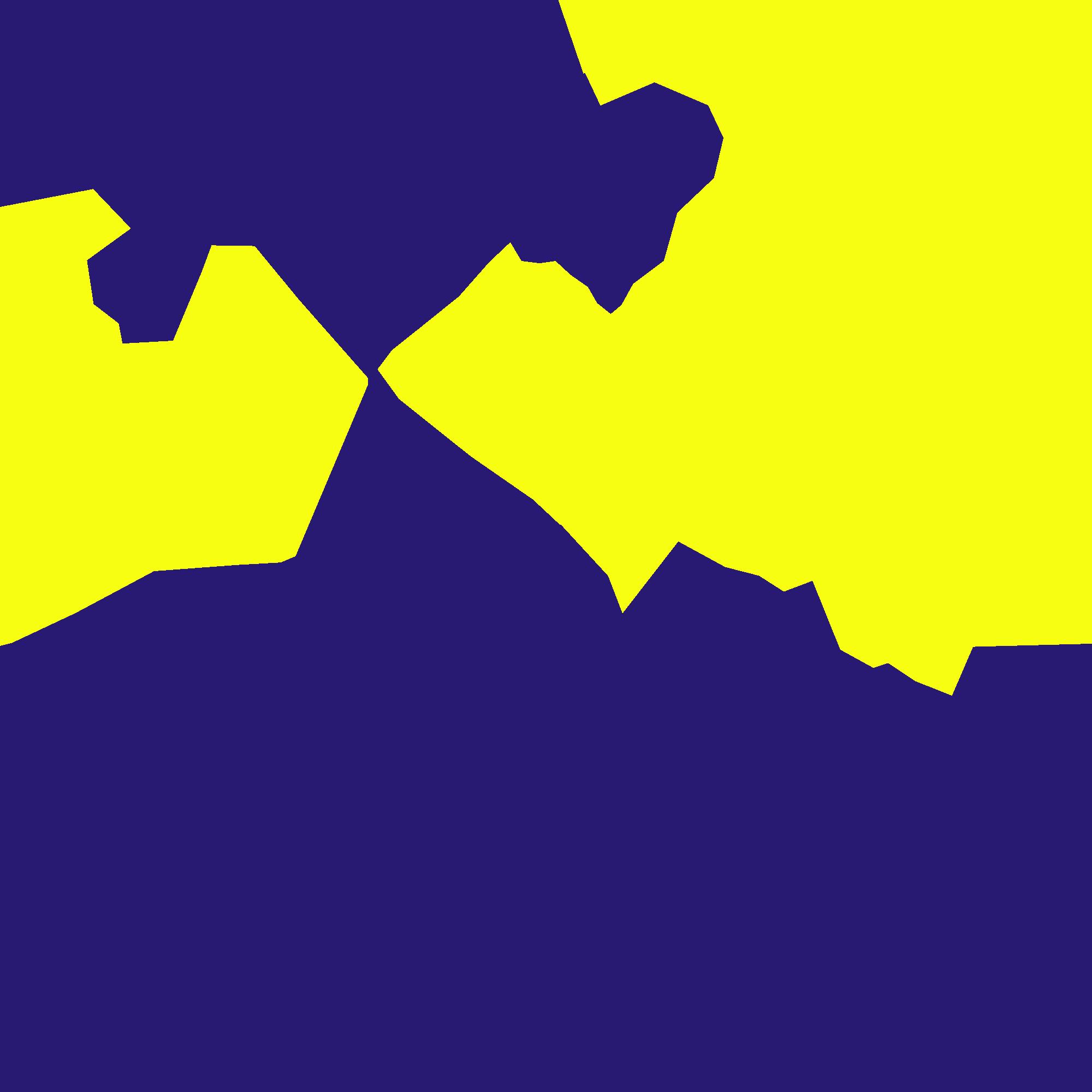}&
\includegraphics[width=4cm]{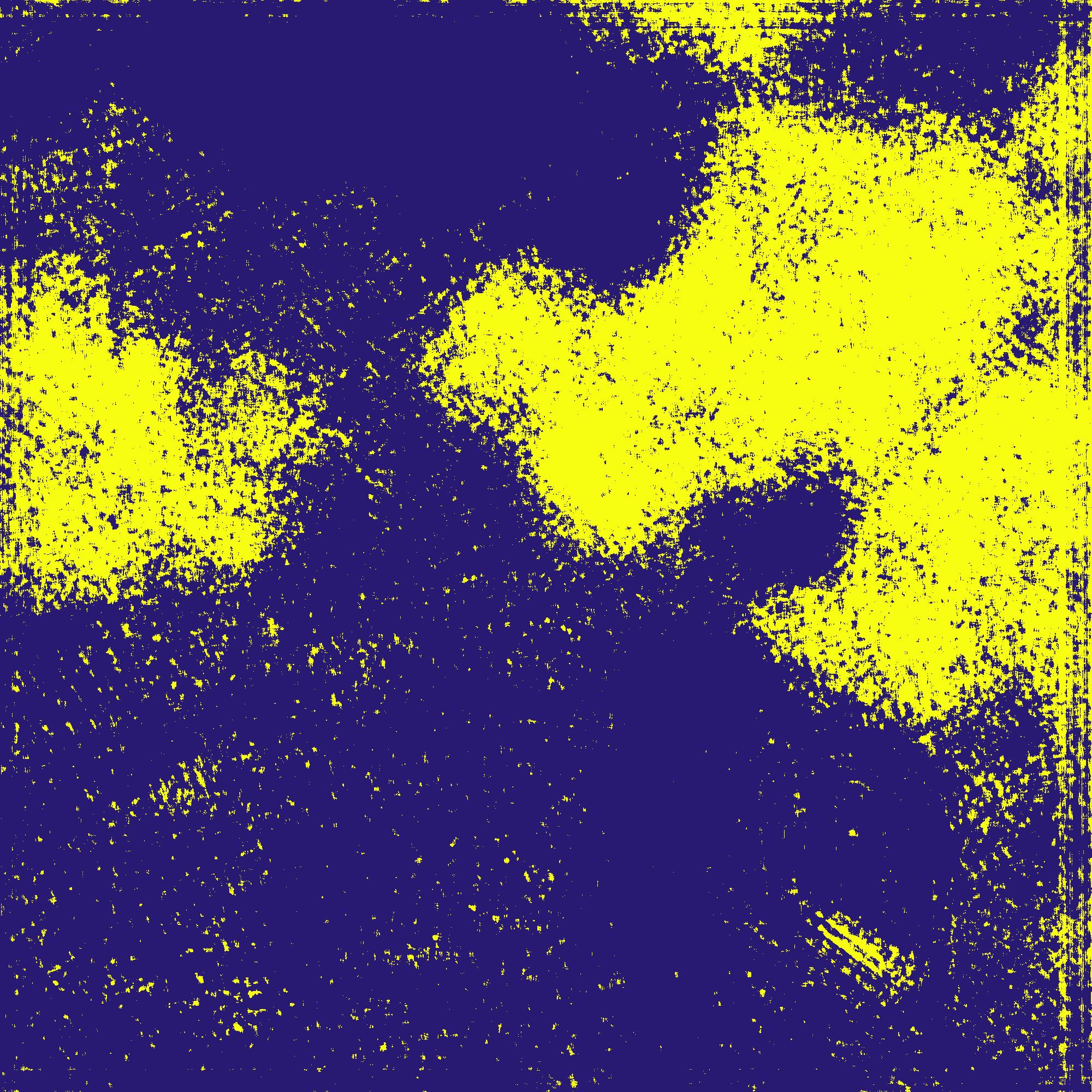}&
\includegraphics[width=4cm]{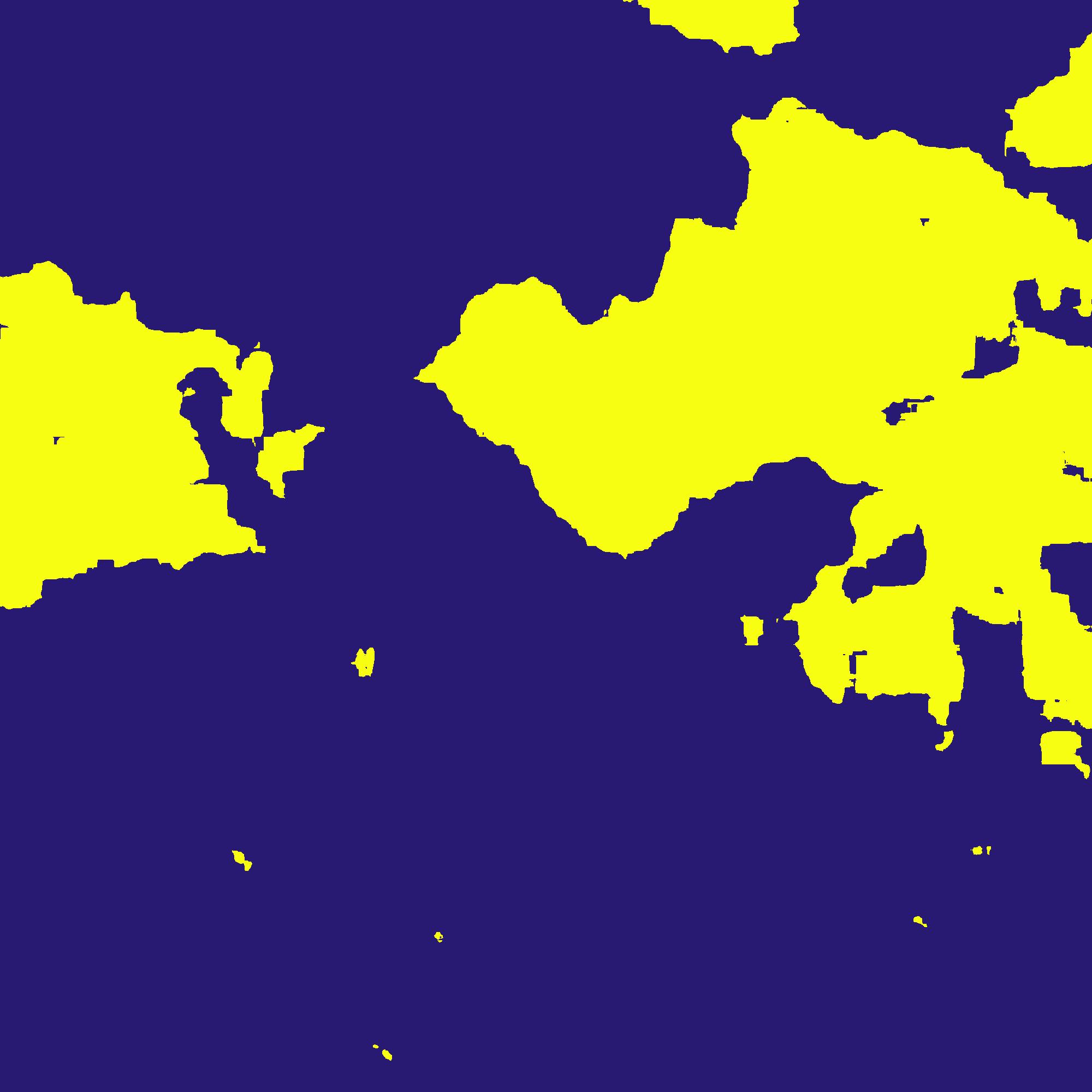}\\
\end{tabular}
\caption{Informal settlement mapping over two test areas in Dar es Salaam, Tanzania, using patch-based CNN and FCN-DK6 \cite{Persello2017DeepImages}. ``informal settlements" in yellow and ``rest" in blue. }
\label{fig:slum_maps}
\end{figure*}

\subsubsection{Revealing socioeconomic inequalities}
The papers mentioned above cast the slum mapping problem as a crisp classification, assuming that a boundary can be drawn to separate formal and deprived settlements. Departing from this dichotomy, Ajami et al. adopt a framework conceptualizing the multi-dimensional nature of deprivation including not only the physical (e.g., poor house material) and financial level (e.g., low-income residents), but also human, social, and contextual variables such as accessibility to healthcare, education, and other services or social exclusion factors \cite{Ajami2019IdentifyingNetworks}. The study introduces a data-driven approach to summarize multiple deprivation variables (both categorical and real-valued) into a single real-valued socioeconomic index, named data-driven index of multiple deprivations. A CNN-based transfer-learning method predicts the socioeconomic index values based on VHR images and geographic information system (GIS) features. The results show that an ensemble non-linear regression model, combining the results of the CNN and models based on hand-crafted and GIS features, can explain 75\% of the variation in the poverty index obtained from household data.

Other works have applied DL models to nighttime satellite images, street-view, and aerial imagery to infer socioeconomic conditions. Jean et al. use a CNN-based model to predict economic well-being across five African countries \cite{Jean2016CombiningPoverty}. The CNN model, pre-trained on ImageNet, is fine-tuned to predict nighttime light intensities (used as a proxy for economic activities) corresponding to input daytime satellite imagery. Finally, the CNN-extracted features, along with survey data, are used as input to a ridge regression algorithm to infer the economic well-being. Social, environmental, and health conditions are extracted in \cite{Suel2019MeasuringImagery} by a  DL method applied to street-view images for major cities in the UK. Abitbol et al. use a modified EfficientNetB0 CNN architecture  \cite{Tan2019EfficientNet:Networks} to predicting socioeconomic status across France from aerial images and use activation maps to interpret the urban topology \cite{Abitbol2020SocioeconomicNetworks}.

\subsection{Deliver tenure security for all} \label{tenure security} 

Secure property rights and efficient registration systems are essential for the modern economy. They give guaranty to individuals and businesses to invest in land, creating the conditions for improving the livelihoods and sustainable management of natural resources, and enabling governments to collect property taxes, which are necessary to finance infrastructure and services to citizens. Unfortunately, a mere 30\% of the global population has legally registered rights to their land and homes \cite{Dale1999LandAdministration, Enemark2014Fit-for-purposeAdministration}, which means that more than 5 billion of the world’s 7.8 billion people do not have documented land rights. Moreover, this percentage drops down to 10\% in African countries \cite{Boone2019LegalAfrica}. The insecurity of land tenure and property rights is often at the root of poverty and inequality \cite{Lakner2019HowPoverty, Conforti2011Looking2050}, leading to legal conflicts, unequal economic systems, locks of assets, challenging effective and democratic governance principles.

\begin{wrapfigure}{r}{4.0cm}
\vspace{-0.5cm}
\begin{mdframed}[backgroundcolor=gray!20] 
{\em ``The insecurity of land tenure and property rights is often at the root of poverty and inequality''}
\end{mdframed}
\vspace{-0.5cm}
\end{wrapfigure}

The 2030 agenda recognizes the fundamental role of land rights security in several targets and indicators under SDG 1, 2, 5, 11, 15, and 16. The correct registration of land tenure rights directly impacts food security, environmental sustainability, and advancing women's empowerment worldwide. In many countries, the land is communally owned, but tenure insecurity is often the product of the government's inability to respond to the technical regularization needs \cite{AldenWily2018CollectiveTrends}. Therefore, significant efforts are needed to formalize land ownership of the poor and vulnerable (target 1.4, indicator 1.4.2). Secure access to land is essential for small-scale agricultural producers to invest in their land and contribute to the market (targets 2.3 and 2.4). It is also fundamental for gender equality, ensuring women's rights to land tenure (SDG 5). The authors in \cite{Meinzen-Dick2019WomensEvidence} show that secure land tenure for women improves investments in agricultural developments and enhances the chance of women involvement in family food and agricultural productivity.
Lack of tenure security also \new{}{impacts} the development of sustainable cities (SDG 11), the management of natural resources (SDG 15), and the synergy between land administration agencies, courts, and legal support services (SDG 16).

Strategies to support these goals rely partly on the development of land administration systems (LAS) to formalize land rights and implement land-related policies \cite{Williamson1997TheCountries}. There is a clear need for innovation for fast, accurate, and cost-effective cadastral mapping needed for LASs \cite{Koeva2020InnovativeMapping}. The traditional surveying methods prove to be quite costly, slow, and labor-intensive. In response, fit-for-purpose (FFP) land administration \cite{Enemark2016GuidingCountries} advises and supports the development of new technologies using remotely sensed data and taking the country context into account. 
The FFP concept is also included in the recently developed \textit{framework for effective land administration} (FELA) developed by the UN expert group on land administration and management, which is acting as a standard at international level \cite{UN-GGIM2019FrameworkOn}. 
Spatial land rights recording, i.e., cadastral mapping, is the most expensive part of a land administration system \cite{Williamson2010LandDevelopment}. Automation or semi-automation of cadastral boundary delineation based on satellite or UAV images has been investigated since physical objects often coincide with visible cadastral boundaries and can be detected through image analysis \cite{Luo2017QuantifyingVanuatu, Crommelinck2016ReviewMapping, Manyoky2012UnmannedApplications, Jazayeri2014AInformation, Koeva2016UsingRwanda}. The extracted boundaries should be combined with legal information, a procedure known as \textit{adjudication}, and should incorporate local knowledge from human operators. 

\begin{figure}[tbp]
\begin{center}
 \includegraphics[width=1\linewidth]{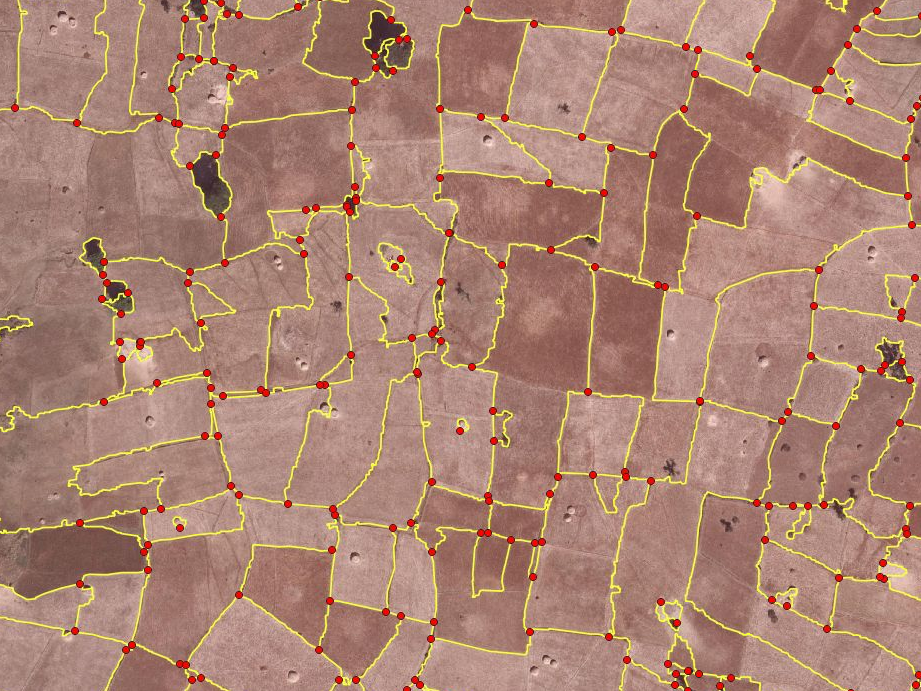}
 \end{center}
 \caption{Cadastral mapping tool (its4land) applied to a rural areas in Ethiopia}
  \label{fig:its4land}
 \end{figure}

Initially explored methods for the automated extraction of cadastral boundaries from EO images are based on image segmentation and edge detection \cite{Crommelinck2016ReviewMapping}. However, the main disadvantages of these methods are sensitivity to intra-parcel variability and dependence of the selected parameters \cite{Garcia-Pedrero2017ASegmentation, dragomir2009review}. 
Better results have been achieved using learning-based contour detectors such as the globalized probability of boundary (gPb) \cite{Arbelaez2011ContourSegmentation}, which combines brightness, color, and texture cues into a globalization framework using spectral clustering.
Recent studies explored DL methods \cite{Crommelinck2019ApplicationImagery, Xia2019}. The consortium of \textit{its4land} \footnotemark{} project developed a suite of open-source solutions for land tenure recording using EO data \cite{Koeva2018Its4landMapping, Koeva2020InnovativeMappingb, Koeva2017TowardsMapping, Koeva2020InnovativeMapping}. They developed methods based on gPb, SLIC superpixels, and CNN to extract cadastral boundaries and a strategy to assign costs to each line incorporating local user knowledge. 
This work resulted in an open-source plugin for QGIS providing user-guided delineation functions calculating least-cost paths along the extracted and weighted boundaries.
Experiments were conducted using aerial images acquired in Ethiopia and UAV images from Rwanda and Kenya (Fig. \ref{fig:its4land}). Overall, the obtained results based on CNN-derived boundaries achieved a precision of 76\%. 
The use of this semi-automated interactive method leads users to spend 38\% less time and 80\% fewer clicks compared to manual delineation \cite{Crommelinck2018InteractiveData, Crommelinck2019ApplicationImagery}. 
Following this research line, Xue et al. \cite{Xia2019} explored the potential of FCNs to extract cadastral boundaries in urban and semi-urban areas in Rwanda using UAV data. The authors adopted the FCN-DK architecture, which compared to gPb and multiresolution segmentation, resulted in overall better performance.
Nevertheless, the performance of any automated method depends on the presence of visible objects delimiting the boundary of the property (e.g., fences, pathways, walls, roads, land cover transitions).
\footnotetext{https://its4land.com/}
In support of the full land recording process, Chipofya et al. developed an approach incorporating hand-drawn sketch maps with remotely sensed data \cite{Chipofya2020LocalLADM}. The approach converts the raster sketch map into vector automatically, and the hand-drawn symbols are detected and recognized using a CNN. The system performs a stroke-based image segmentation wherein boundaries of sketched objects are drawn and delineated. Finally, the concepts corresponding to the detected symbols are applied to the image segments based on distance and a fixed set of rules specifying spatial constraints on configurations of different types of features.

\subsection{Climate Action}

{The current scenario of climate change and the projections from climate models call for a definite and urgent action as requested by SDG 13 \cite{romm2018climate, archer2010climate}. Extreme events are more severe, more frequent, and more unexpected in space and time \cite{field2012managing}. The Earth system is actually changing globally but also at local and regional scales, with huge implications in ecosystems, biodiversity and agriculture just to name a few. 
In this scenario, humanity faces the challenge of both mitigation and adaptation to climate change, that is; try to reduce emissions as much as possible while preparing for unavoidable consequences that are no longer a future, but a reality \cite{rolnick2019tackling}. Machine learning and DL, in particular, can help in the myriad of aspects concerned in both issues. The 2030 agenda focuses mainly on adaptation aspects with target 13.1 demanding to strengthen resilience and adaptive capacity to climate-related hazards and natural disasters in all countries. }

{Mitigation of greenhouse gas (GHG) emissions requires important changes to electricity systems, transportation, buildings, industry, and land use. Adaptation requires planning for resilience and disaster management, given an understanding of climate and extreme events, see \cite{rolnick2019tackling} for a organized collective effort to synthesize both the methods and challenges.}

\subsubsection{DL for climate change mitigation}

{Reducing emissions can be achieved with machine learning and DL models. 
For instance, several DL models have been used to forecast electricity supply and demand, e.g., create short and medium-term forecasts of solar power \cite{mathe2020pvnet} and wind power \cite{foley2012current}, \cite{deepmind2019machine}
or even to use deep networks to produce demand forecasts that optimize for electricity scheduling costs rather than forecast accuracy \cite{donti2017task}. DL in combination with RS satellite imagery has been also used to generate size and location data for rooftop solar panels \cite{malof2016automatic, yu2018deepsolar}, and there are some deep networks that estimate the state of the system \cite{jiang2016short, pertl2016voltage}. 
As electricity gets transported from generators to consumers, some of it gets lost as resistive heat on electricity lines. Prior work has performed predictive maintenance using LSTMs \cite{bhattacharya2017deep} and neural network-plus-clustering techniques \cite{nguyen2018automatic} on electric grid data. 
Another important field of action is that of transportation. Decarbonizing transport is essential to a low-carbon society, and there are numerous applications where machine learning can make an impact. For instance, vehicles can be detected in VHR images accurately \cite{jiang2015deep, mundhenk2016large, deng2017toward} and image counts can serve to estimate average vehicle traffic \cite{kaack2019truck}. Neural networks have also been used for analyzing preferences of customers traveling by high-speed trains.  
Many critical systems inside buildings can be made radically more efficient. Deep autoencoders can be used to simplify information about machine operation so that deep neural networks can then more easily predict multiple kinds of faults \cite{jia2016deep}. Occupancy detection in buildings can help identify energy demands, a problem where deep neural networks have been also applied \cite{zou2018towards}. DL can also help to monitor and optimize the operations in smart buildings \cite{ateeq2019multi}. 
Machine learning may be able to help with many aspects of CO$_2$ sequestration. While still in its infancy, we have seen recent approaches on the use of convolutional image-to-image regression techniques for uncertainty quantification in a global carbon storage simulation study \cite{mo2019deep}. Such models can help in the development of novel strategies to monitor and develop underground carbon sequestration techniques.
}

\subsubsection{DL for climate change adaptation}

{
We use general circulation and Earth system models to anticipate climate scenarios on our planet, and to inform local and national governments for decision making. 
Models  have become very precise in projecting scenarios, but still they disagree in some particular cases and are very computationally expensive to run. 
Machine learning in general and DL in particular can help to mitigate both aspects.
The largest part of the uncertainty comes from the parameterization of clouds and aerosols in the models, which have clear implications as bright clouds block sunlight and cool the Earth. Deep neural networks have been used to emulate the behavior of high-resolution cloud resolving simulations at a fraction of the computational cost \cite{Gentine2018}. Improvements are expected by the combination of DL and process understanding in a new form of hybrid modeling approaches that are data-driven while respecting the fundamental laws of physics  \cite{Reichstein2019,CampsValls21wiley}. 
Future improvements in climate modelling will necessarily have to account for the proper characterization and modeling of ice sheet dynamics and sea level rise, yet machine learning has not approached such problems yet systematically \cite{bolibar2020deep,Nieves21sr,CampsValls21wiley}.

Weather models are optimized to track the rapid, chaotic changes of the atmosphere, and DL has recently impacted the associated problems. For instance, deep networks are now heavily used to make local forecasts from coarse 10--100 km climate or weather model predictions~\cite{Li2019}, while other researchers try to translate high-resolution climate forecasts into risk scenarios, e.g. of localized flooding patterns from past data \cite{perignon2018patterns}, which has clear impacts on individuals. 
Accurately forecasting hazards and their impacts has societal, economical, and environmental implications. DL is now present in initiatives involving preserving ecosystems at risk \cite{bragilevsky2017deep}, monitoring the risk of food insecurity \cite{wang2018deep} and deployment of a swift, effective disaster response \cite{voigt2007satellite}. 
Yet, humans can also intervene in the system directly. This is the field of geoengineering. For example, neural networks approaches could facilitate the fast release of aerosols in both space and time \cite{de2019stratospheric}. Modeling impacts is also of high relevance; the authors in \cite{di2016assessing} use deep neural networks to estimate the effects of aerosols on human health, while Crane-Droesch et al. use them to estimate the effects of solar geoengineering on agriculture \cite{crane2018using}. 
Finally, we should note that geoengineering raises many ethical questions, where explainable, accountable AI and fair learning should \new{}{be part of the discussion.} 
}


\subsection{Life on Land}
SDG 15 aims to protect, restore and promote sustainable use of terrestrial ecosystems, sustainably manage forests, combat desertification, halt and reverse land degradation, and stop the loss of biodiversity. Achieving this goal has far reaching consequences that are closely interlinked with many other SDGs. We review here several application domains where DL and EO play a central role.

\subsubsection{Sustainable forest management}
{The relevance of sustainable management of forests is linked directly to SDG 15, and target 15.2 in particular, but goes well beyond that. Indeed, the “State of the World’s Forests 2018” report of the Food and Agriculture Organization (FAO) \cite{rhFAOreport2018} identifies that forests and trees are relevant for 28 targets from ten different SDGs. On the one hand, forests are a key variable to mitigate the effects of climate change, they protect soil and water, and contain more than 75\% of the world’s terrestrial biodiversity. On the other hand, forests provide products and services such as food, medicine, and fuel that are of high socio-economic importance in particular in rural areas. The combination of RS and DL has been extensively used to monitor forests, e.g. by producing global forests maps \cite{rhMazza2019}, delineating individual tree crowns in aerial imagery \cite{rhWeinstein2019,rhDong2019}, or performing damage assessment after storms \cite{rhHamdi2019}.}

Despite their relevance, the loss of the world’s forests through deforestation and forest degradation is an increasing issue destroying natural habitats, limiting resources for the world’s poorest and in the long term worsening climate change by significantly contributing to CO$_2$ emissions. Reasons for deforestation include tree logging for materials, mining, and farming. Agriculture producing palm oil, beef, soy, pulp and paper is responsible for nearly three quarter of tropical deforestation \cite{rhFAOreport2018}.
\begin{wrapfigure}{l}{4.0cm}
\vspace{-0.5cm}
\begin{mdframed}[backgroundcolor=gray!20] 
{\em ``Remote sensing is the ideal tool to monitor large forests that are difficult to access''}
\end{mdframed}
\vspace{-0.5cm}
\end{wrapfigure}
Deforestation is mostly happening in rural areas and often performed in secrecy, which requires monitoring large regions that are difficult to access. Thus, RS is the ideal tool to map and monitor forest that inspired the publication of open data (e.g., in the context of the ``Understanding the Amazon from Space'' challenge organized by Planet \cite{rhPlanetChallenge2017}) as well as the usage of DL approaches. Several works focus on experimental comparison between different network models and shallow learners \cite{rhLee2020,rhdeBem2020} or apply ensembles of different CNN architectures (e.g., \cite{rhBragilevsky2017} - a participant in the aforementioned challenge). In \cite{rhMaretto2020}, deforestation mapping is modelled via spatio-temporal deep CNNs by taking several domain-specific components (e.g. handling of clouds) into account. Modern approaches go beyond a mere mapping of forest areas or directly deforestation and instead aim for identifying possible reasons for forest loss. ForestNet \cite{rhIrvin2020} not only proposes a deep convolutional network to characterize the processes leading to deforestation but also provides a data set based on Landsat 8 imagery of forest loss events annotated by expert interpreters.

\subsubsection{Wildfire risk}
Wildfires, as one of the major factors contributing to deforestation, are becoming more frequent and destructive due to several reasons including higher temperatures, increased droughts, fuel accumulation and dead vegetation, as well as increased population density in close proximity to forests and wildlands.
Traditionally, wildfires are detected by human observers either by chance (and then reported to local emergency numbers) or from dedicated watchtowers. Current works aim to complement or even replace the latter by deploying ground-based camera networks (e.g., HPWREN and Alert Wildfire in the State of California, USA, where in 2018 more than 8,000 wildfires burnt 800,000 hectares of land).
The camera feed of these networks can then be automatically analyzed by DL approaches as e.g., in \cite{rhGovil2020} which uses an Inception network to detect smoke. An alternative is the usage of UAVs as proposed in \cite{rhZhao2018}, which uses a saliency-based system to generate image region proposals that are then analyzed by a standard CNN for classification. Satellite imagery can be used for wildfire detection as well but comes with its own challenges. Geosynchronous satellites such as GOES~16 or GOES~17 constantly observe large parts of a hemisphere but have a rather coarse resolution of several square kilometers which makes detection of wildfires in their early stages difficult. Nevertheless, their image data has been used in combination with DL for wildfire detection (e.g., in \cite{rhToan2019}). Orbiting satellites such as MODIS, VIIRS, Landsat, and Sentinel-1/2, on the other hand, have a much finer spatial resolution but revisit times of several hours to days. In particular, SAR sensors offer unique benefits as they are able to penetrate clouds and smoke and are independent of daylight. Sentinel-1 time series data and DL have for example been used in \cite{rhBan2020} to provide a near real-time progression monitoring of wildfires. 

Beyond detection and monitoring of wildfires, forecasting their future burns and spread is another important application area \cite{rhRadke2019, Bergado2021PredictingLearning}. Bergado et al. \cite{Bergado2021PredictingLearning}, use a big geodata set to predict wildfire burns. They design FCNs for predicting daily maps of the probability of a wildfire burn over the next 7 days utilizing an extensive set of wildfire related input variables taken from various data sources. 29 quantitative features are selected as input to the models. These features encode factors associated to wildfire burn such as topography (elevation, slope, and aspect), weather (temperature, humidity, solar radiation, rainfall, wind speed and direction, and lightning flash density), proximity to anthropogenic interfaces (distance to roads, distance to power lines) and fuel characteristics (fuel type, fuel moisture, emissivity). Historical wildfire burn records for Victoria, Australia, collected over the period of 2006–2017 are used for training and testing the DL models. DL and RS have also been used for post-event analysis e.g., for damage assessment \cite{rhFarasin2020} or to analyse the impact of wildfires on tree species \cite{rhZhou2020}.

\subsubsection{Bio-physical parameter estimation}
{The problem of retrieving bio-geophysical parameters spans a wide variety of applications and has found a direct impact on achieving the SDGs. The related goals require creating spatially explicit and temporally-resolved maps of quantities and essential climate variables 
to monitor vegetation status and health, agricultural and forest production. Parameters should be estimated in a consistent and standardized manner to improve accountability. The use and abuse of vegetation indices as proxies of vegetation status and health has been challenged recently by machine learning approaches, from nonlinear generalization of indices \cite{CampsValls21sciadv} to more advanced, yet supervised, machine learning models \cite{CampsValls21wiley}.  
Many parameters are now estimated using machine learning; for example, surface temperature and moisture are key parameters for weather prediction with great impact on agriculture and environment, in ecology, hydrology, meteorology, and biology, while leaf area index (LAI) and fractional vegetation cover help in assessing the vegetation cover and dynamics, with implications on crop production. Several seminal works relevant for DL parameter retrieval have been published with focus on Earth Sciences \cite{Reichstein2019,CampsValls21wiley}, environmental applications \cite{yuan2020deep}, and remote sensing \cite{zhu2017deep}.}

{Land parameter retrieval often concerns bio-chemical parameters but can also include physical parameters such as land surface temperature (LST), which was retrieved from microwave radiometer data with DL in \cite{tan2019deep}, which was tested on reference data from both ground stations and other optical satellite data with good results. Leaf-Area-Index (LAI) and Leaf-Chlorophyll-Content (LCC) have been retrieved with optical sensors and using neural networks, yet mostly using shallow architectures \cite{verrelst2015optical}.}

{Retrieving parameters can be often hampered by the scarcity in measurements and observations to spatialize them with machine learning. This is the case of relevant parameters for monitoring the land and vegetation such as canopy water content (CWC). For such a case, one can resort to radiative transfer models to generate a look-up-table of expressive simulations to learn from and upscale it in the Google Earth Engine globally \cite{Haro19cwc}. There are some other cases where samples are available in big databases but not sufficiently complete, with many missing attributes or uncertainty in the wild. This was the case of important leaf and plant traits, like phosphorus or nitrogen concentrations, that were not upscaled until the exploitation of the TRY database along with multi-sensor fusion and machine learning \cite{Moreno18}. Upscaling carbon, heat and energy fluxes from eddy covariance data has been recently tackled with all kind of machine learning models and neural networks in particular. The key parameters for sensing the health and sensitivity of our warming planet are gross primary production (GPP) and net ecosystem exchange (NEE). Their estimation with neural networks and ensemble methods allow us to quantify global land-atmosphere interactions and benchmark land surface model simulations \cite{Jung18fluxcom,Jung19flux}.}

{Research in farming applications also relates to biological parameter retrieval. Often though, the goal is not to use predictions as parameters in models, but as proxies for the health condition of crops in so-called smart farming applications. By monitoring and optimising these vegetation indices the goal is to increase crop yield. The variables of interest such as crop type, crop yield, soil moisture and weather variables, can also be used to model and understand the ecosystems that farming effects \cite{kamilaris2018deep}. Most often though, it is applied to data sets covering only smaller regions of agricultural areas. As opposed to biological parameter retrieval applications, DL is frequently used in farming applications. Some country level work on agriculture has been done for e.g. corn crop yield, \cite{kuwata2016estimating} and wheat \cite{Ola19xai}, but little work exists on larger scale studies where predictions could be used in models. The authors in \cite{kim2017massive} provide a comparison of several artificial intelligence methods on a case study in mid-western USA.}

Forest cover, biomass and vegetation height are other types of biological parameters which are of high importance to understand and monitor the Earth, with obvious societal and economical implications. DL has also been applied to this problem although mostly on continental level scale, e.g. in \cite{ye2019projecting} used the LSTM networks, while the authors in \cite{khan2017forest} modeled forest dynamics over a 28-year period by stacking time-series and formulating the task as a change-classification problem, and \cite{zhang2019deep} predicted above ground forest biomass from Lidar and Landsat 8 data with Stacked Sparse Autoencoders (SSAE). The authors in \cite{lang2019} map vegetation height densely at 10-meter resolution from stacks of Sentinel-2 multi-spectral optical satellite imagery at country-scale using CNNs with a regression loss.

\subsubsection{Wildlife conservation}
Global loss of biodiversity is observed at all levels~\cite{Ceb20}, and mammals are no exception, with one fifth of them at risk of extinction~\cite{You16}. Conservation relies heavily on monitoring to estimate biodiversity, as well as resources to sustain life and risks related to human activities (hunting, poaching, expanding agriculture, etc). Despite the urgency of protecting animal populations, the population monitoring is more often done locally in reserves, by experts on foot, and hardly meets the scaling and update requirements to monitor fauna effectively. Due to their larger field of view and the relatively high revisit time potential, satellites \cite{duporge2020using} and -- more importantly -- drones \cite{rey2017detecting}, are considered more and more for surveying by wildlife ecologists~\cite{Lin15,Fus20}. Drones open perspectives for monitoring on demand, safe detection of poachers and estimation of grazing potential. To process the sheer amount of data collected by drones, researchers are starting to resort to DL massively, to detect animals in the wild with object detection pipelines~\cite{Kel18,norouzzadeh2018automatically,Eik19,Pen20}.

\begin{wrapfigure}{r}{4.0cm}
\vspace{-0.5cm}
\begin{mdframed}[backgroundcolor=gray!20] 
{\em ``Remote sensing and deep learning can accelerate conservation efforts and play a central role in the battle against poaching''}
\end{mdframed}
\vspace{-0.5cm}
\end{wrapfigure}
\begin{figure}[!t]
    \centering
    \begin{tabular}{c}
        \includegraphics[width=\columnwidth]{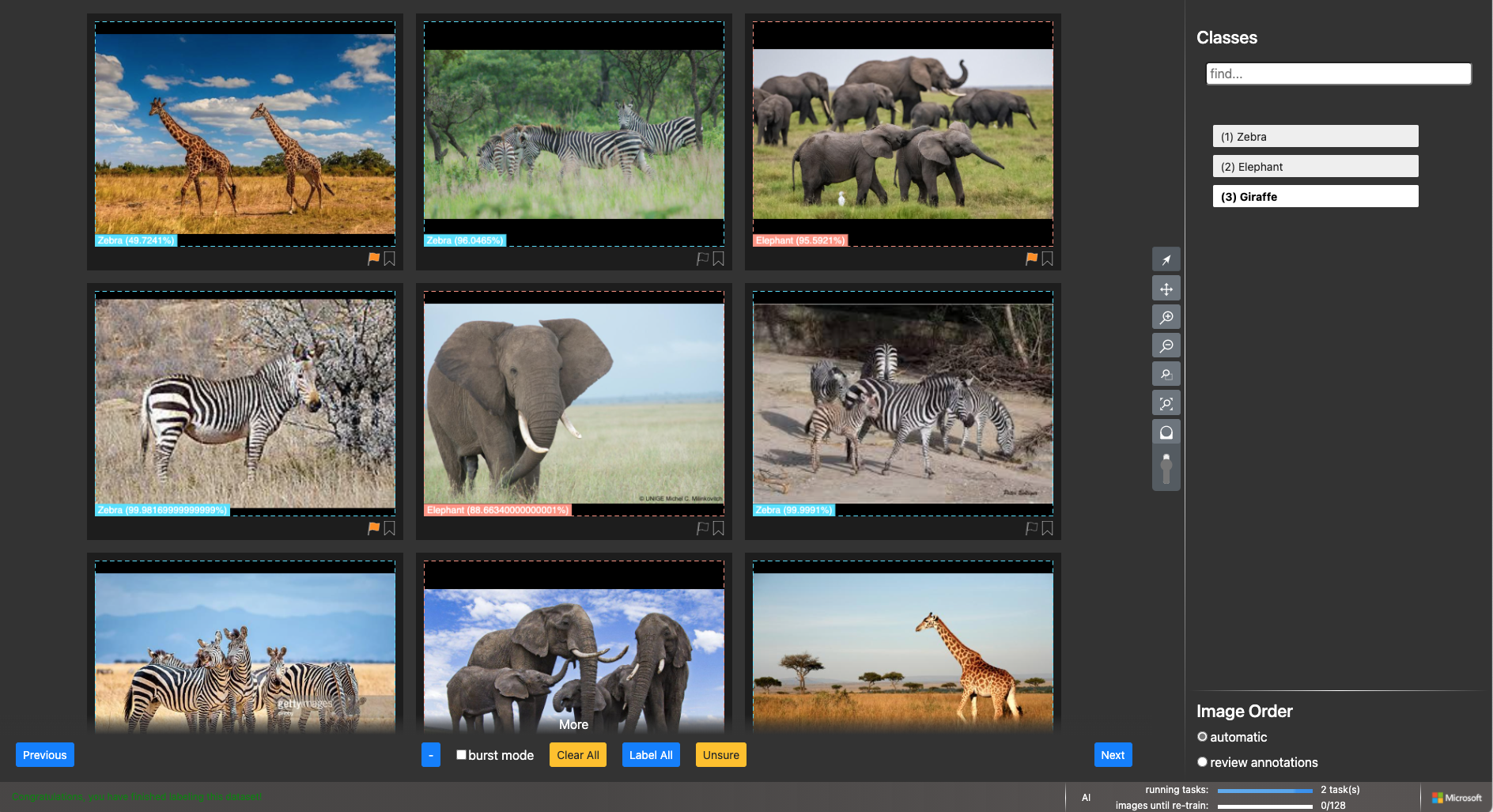}\\
        a) Classification of camera traps images\\
        \includegraphics[width=\columnwidth]{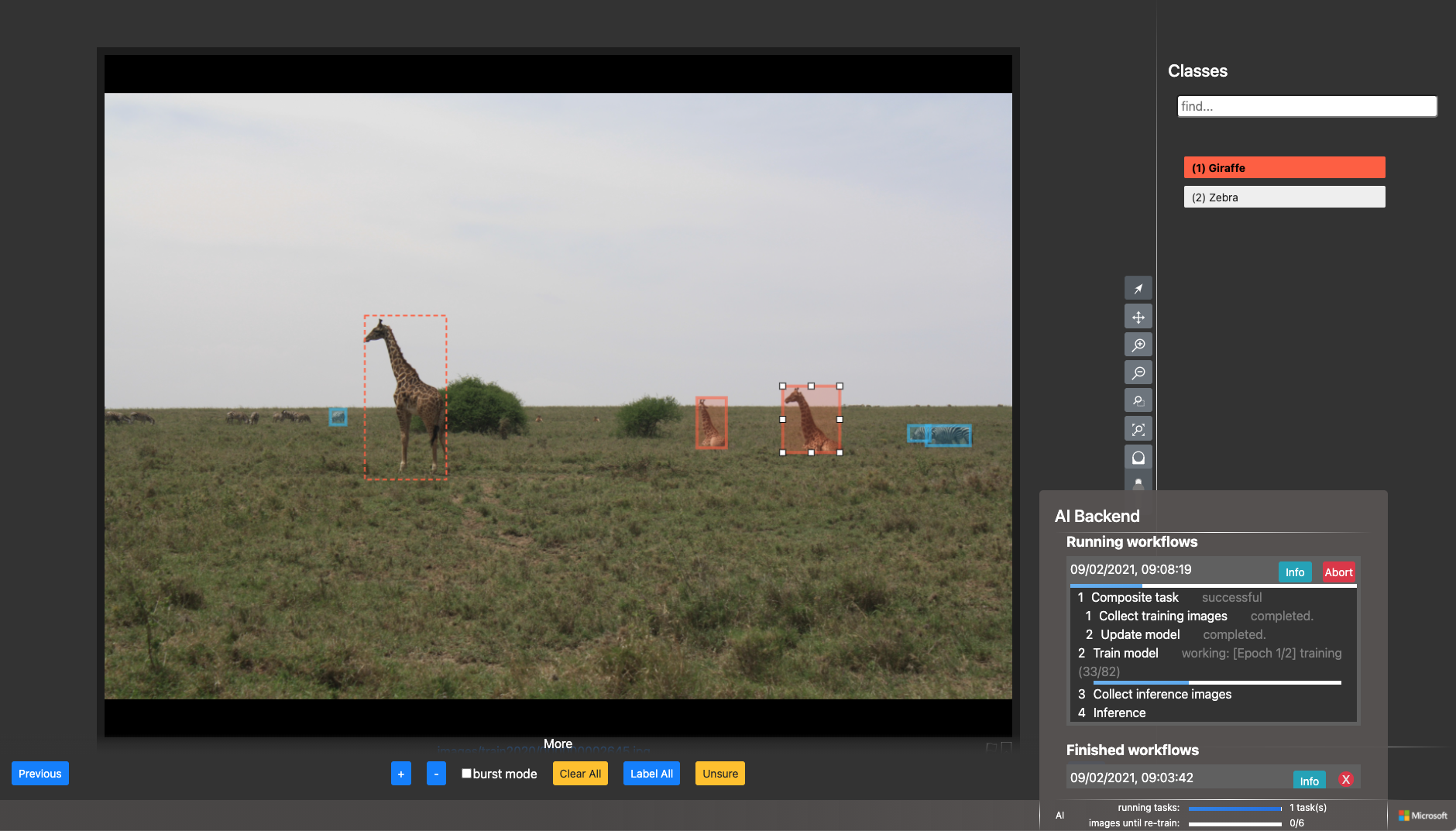}\\
        b) Detection in community images\\
        \includegraphics[width=\columnwidth]{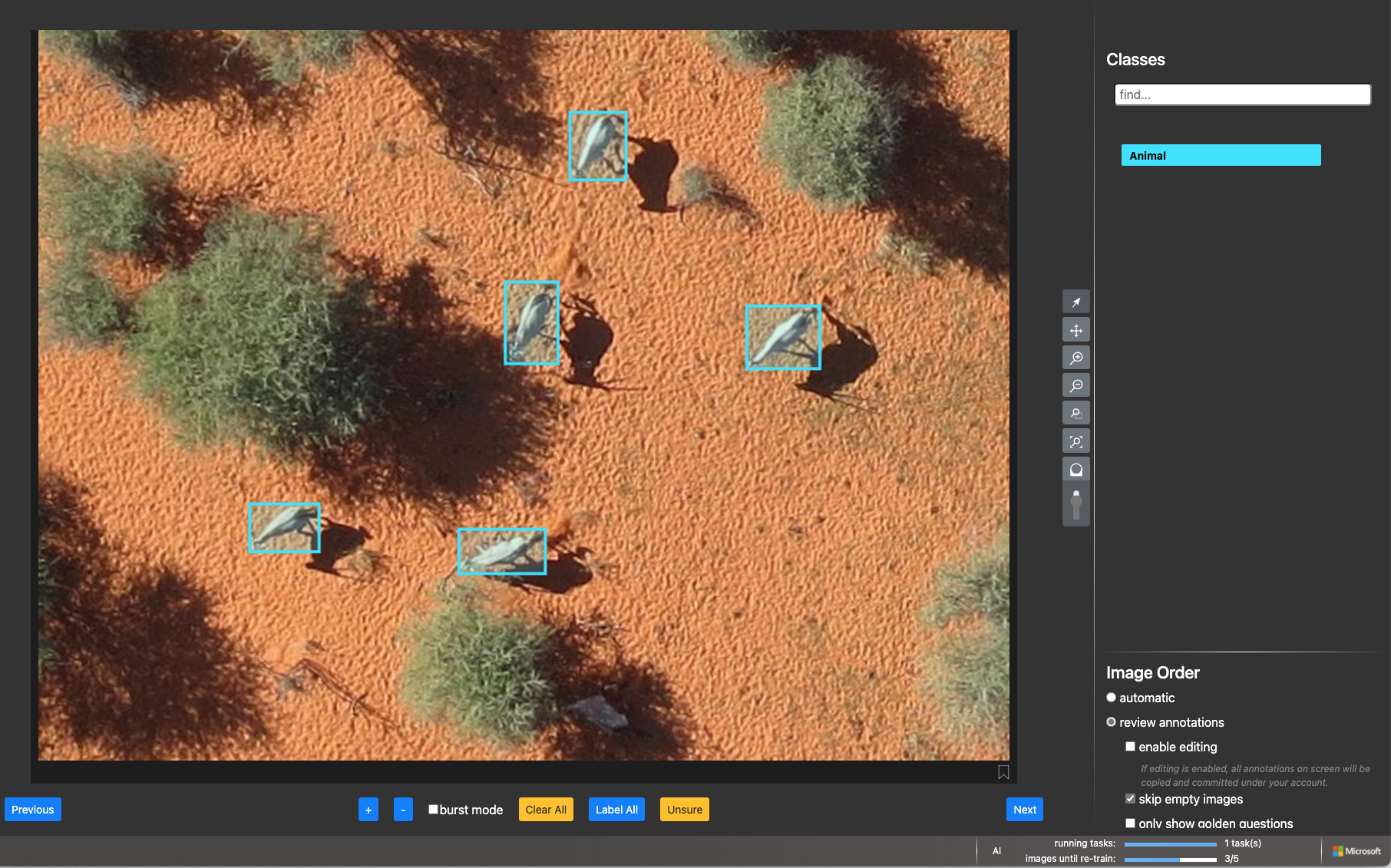}\\
        c) Detection in UAV data
    \end{tabular}
    \caption{Three deployment cases of the AIDE platform~\cite{Kel20} in camera traps images, on (a) camera traps image classification; (b) on animal detection from tourists and photographers pictures acquired during a safari (using the `Great Zebra and Giraffe Count' (GZGC) campaign (http://lila.science/datasets/great-zebra-giraffe-id)) and (c) detecting wildlife in UAV images in the Kuzikus reserve in Namibia. In the GZGC case, a typical deployment is shown: a user is editing the detections, while a model is training in the cloud (bottom right box); the predictions of the current model (dashed lines) are also used as a guidance. }
    \label{fig:wild}
\end{figure}

These efforts go hand in hand with computer vision-based community efforts aiming at processing wide archives of camera traps images, i.e. static cameras placed at strategic locations in reserves~\cite{beery2019efficient}. To support research in deep learning-based animal conservation, a number of software suites are being proposed, including AIDE~\cite{Kel20}, which allows ecologists to upload their camera traps or aerial survey and to deploy pre-trained (or own) models in the cloud on Microsoft Azure. Examples of deployments of AIDE in camera traps image classification and single animal  detection are reported in Figure~\ref{fig:wild}. The questions of accuracy with respect to animal size and image resolution, or when related to the ratio of background vs animal occupancy in the images (the animals only occupy a fraction of the data collected) are central in these papers. Detection and tracking of poachers is also on the rise, with approaches using thermal images at night~\cite{Les19} or based on deep reinforcement learning~\cite{Wan19}.

{In the context of animal censuses, a question of wide interest is the time efficiency versus the number of animals these algorithms miss. Precise counts are of prime importance and a low recall would force rangers to go through the entire image collection for verification, which would negate the benefits of the DL detection pipeline. Recent research compared DL and citizen science counting methods~\cite{Tor19} and concluded that both led to similar accuracy, with a significant speedup achieved when using DL. Also, an active research field is the joining of these two worlds via active learning~\cite{Set12} algorithms: by allowing interactive back and forth between the annotators and the DL models, significant speedups, as well as increased generalization to new campaigns and transfer to new reserves, has been achieved~\cite{Kel19}. Finally, these interactive pipelines are nowadays made accessible to the large ecological community, for instance via web-based platforms enabling interactive annotation guided by DL models classifying or detecting in the background~\cite{Kel20}.}

\section{Challenges and future opportunities}
The previous section shows several examples of geospatial applications where DL and EO allow a systematic investigation of global phenomena, providing continuous and spatially consistent information supporting evidence-based decision-making and local interventions. We expect that the coming decade will see a surge of research in this direction, with innovative methodological developments and with an increase in the number and scope of applications in support of the SDGs. However, several questions remain to be addressed: some are purely scientific, others are at the interface between scientific communities, stakeholders and decision-makers. Will the EO scientific community succeed in producing accurate, reliable, consistent and up-to-date geospatial information? Moreover, are these results trustworthy for governmental authorities, stakeholders and local communities? In other words, are DL models trusted by non-experts, who are in charge of decision-making and policy development? The success of evidence-based decision making largely depends on the trust that people have in the data. Transparent data analysis methods and clear communication are fundamental to set proper expectations and build trust between data providers and decision-makers.

\subsection{Open challenges}

\subsubsection{Uncertainty quantification} To be of true value and gain trust by people, DL models need to provide an indication of the reliability of model predictions. Assigning well-calibrated uncertainties to model outputs plays a critical role in many real-world applications. A significant additional benefit of uncertainty estimates assigned to each data point of the model output is that this creates a practical interface to more traditional post-processing steps using Bayesian models at their core. We can define uncertainty within DL in a twofold way: inherent to all models is epistemic uncertainty, inherent to all data is aleatoric uncertainty. The former captures the dissimilarity of unseen data compared to what our model has been trained on, i.e. samples that lie within the training distribution have low epistemic uncertainty, samples that are out-of-distribution (OOD) have high epistemic uncertainty. Aleatoric uncertainty results from noise inherent in the observations, such as sensor noise \cite{kendall2017uncertainties}. 
In addition, there are uncertainties in the spatial domain that arise, for instance, when variables are aggregated over spatial units (e.g., districts or administrative units), resulting in the so-called modifiable areal unit problem (MAUP). It is thus important to consider all sources of uncertainties and their propagation through the whole processing pipeline that affect the quality of the final product.  

\subsubsection{Data quality quantification for decision making}
It is essential to realize that to support evidence-based policy-making and promote data-driven policy and decision-making, the quality of data products must be carefully assessed and carefully communicated so that non-experts can understand. On the one hand, we encourage the scientific community to pay more attention to how data quality is assessed and communicated. On the other hand, we promote policy developers to incorporate data uncertainties in the decision-making process explicitly. In this respect, we recognize the importance of defining standard data quality measures. Thus, we promote a tighter collaboration between the scientific community and policy makers to define standards on quality measures to quantify the SDG indicators.

\subsubsection{Model explainability} In addition to uncertainty quantification, there is a growing interest in making machine learning and DL models more interpretable and understandable, aiming at neural networks that provide human-understandable justifications for their output, leading to insights about the inner workings \cite{Chakraborty2018InterpretabilityResults, Samek2020TowardBeyond}. In EO, explainable artificial intelligence is a relatively new field but is quickly becoming important due to the implications that trustable black-box models can have on the usage of DL in societal applications. In the context of agricultural EO, Campos-Taberner et al. investigated how to deepen the understanding of a recurrent neural network for land use classification based on Sentinel‑2 time series \cite{campos2020understanding}. In~\cite{Lev21b}, authors studied how land use can be used to explain automatic prediction of the landscapes scenic value (a form of cultural ecosystem service). To do so, they used semantic bottlenecks \cite{Mar20} as intermediate layers of a regression network predicting landscape beauty from Sentinel-2 images. Forcing the network to choose among human-interpretable solutions, then recombined linearly, the model allows understanding why (in terms of land use) the model predicts a given score.


\subsubsection{Model transferability} One of the greatest challenges of DL in EO is the often limited model transferability. For example, a slum mapping DL model trained in Dar es Salaam is unlikely to produce accurate results in Bangalore or São Paulo. This happens not only because the RS images may be affected by different acquisition and radiometric conditions but also because cities in various parts of the world have different characteristics and definition of what constitutes a slum.
Despite several studies in domain adaptation and transfer learning \cite{Tuia2016DomainAdvances}, model transferability remains a challenge to ensure spatial and national consistency of indicators derived from DL models. Moreover, non-experts might be unaware of this problem. It is therefore essential that DL model developers provide clear guidance to users regarding the domain where the model is expected to produce valid results. 

\subsubsection{Interdisciplinary approach} Addressing global societal problems requires a vast palette of expertise ranging from RS, DL algorithms development, advanced computational skills, as well as domain knowledge in fields such as agriculture, forestry, ecology, urban management and planning, social sciences, land administration, animal conservation, etc.  It requires researchers to collaborate and co-design solutions together with other scientists and engage with stakeholders, industrial partners, local communities, governmental and non-governmental organizations. Barriers between different scientific (and non-scientific) communities are often a challenge for an effective interdisciplinary approach.


\subsection{Future opportunities}


In this paper, we recognize the importance of the availability of data and computational facilities for the success of DL models. Developments in this direction are offering new opportunities to the EO community.
{For half a century, Earth has been under continuous observation by satellites to monitor and understand environmental processes. However, historically, RS data was foremost available to those governmental agencies, research institutes, and commercial companies that had direct access to the corresponding sensors. A mixture of different political, organizational, and legal reasons made a free distribution of acquired data difficult to impossible, resulting in a limited number of mostly small data sets \cite{rhDoldirina2015}. The traditional approach to developing and testing new methods on small and local data sets prevails until today. In particular, in the context of DL, this is problematic as approaches are evaluated on data sets (often consisting of only a single, small image as, e.g., the HSI data set Indian Pines) that do not provide a sufficient amount of independent test samples.}

\subsubsection{Big and open geodata}
{Fortunately, during the last years, large parts of the community have been moving away from closed data and embraced open science principles such as FAIR (findable, accessible, interoperable and reusable \cite{rhStall2019}) and FOSS (free and open-source software \cite{rhFeller2005}). These developments enable transparent and reproducible scientific research, allow the distribution and reuse of data and methods, and lead to more efficient creation of new data products. This led to open code libraries (such as Open RS \url{http://openremotesensing.net}) or the IEEE RS Code Library \url{http://www.grss-ieee.org/publication-category/rscl}), public evaluation servers (such as the IEEE GRSS Data and Algorithm Standard Evaluation website \url{http://dase.grss-ieee.org}), as well as modern benchmark data sets for a multitude of combinations of EO sensors and tasks which are rapidly replacing older small-scale data sets. These are more in line with the actual situation in RS as current EO data has passed the petabyte-scale and poses common big data challenges regarding volume (i.e., the amount of data), velocity (i.e., the temporal pace with which new data is acquired), and variety (i.e., heterogeneity regarding image acquisition such as sensors types and modes, environmental factors) \cite{rhBoulton2018}.

\subsubsection{Cloud computing infrastructures}
New opportunities are also arising from the availability of cloud computing infrastructures that allow to visualize and analyze large-scale data (e.g., Landsat \cite{rhWoodcock2008} and Sentinel data, https://scihub.copernicus.eu/) directly in the cloud without the need for local download, storing, and processing. Examples include Digital Earth Australia \cite{rhDhu2017}, Earth System Data Lab (ESDL) \cite{Mahecha19esdc}, 
the Swiss Data Cube \cite{rhGiuliani2017}, the Copernicus Data and Information Access Services (DIAS) and the Google Earth Engine \cite{rhGorelick2017}. The availability of large and open data sets in combination with powerful computing infrastructures set the premise for researchers to work more cohesively on addressing the environmental and societal challenges of our time.} 

\subsubsection{\new{}{A global picture of global phenomena}} \new{}{Finally, we want to remark that the combination of DL and EO offers the opportunity to obtain a truly global picture of environmental and societal phenomena that go beyond national boundaries as opposed to the data typically collected by national statistical agencies. As discussed in \cite{Scown2020TheGeoscience}, the national-level reporting structure of the SDGs limits the ability to capture environmental phenomena that cross national borders. Moreover, differences in data collection practices of the national offices often result in inconsistent data. EO can provide spatially and temporally consistent data, while DL offers the tools to extract semantic information in an objective and reproducible manner. We thus advocate the use of DL and EO to monitor the progress towards the SDGs and encourage the geoscience and remote sensing community to play an active role in the discussion with stakeholders and policymakers.}







\section{Conclusion}
We have reviewed the latest developments in the context of DL for EO and a large number of applications that contribute to the UN agenda for sustainable development. The combination of DL and EO appears to be a strategic asset that can play an essential role in addressing many of the challenges raised by the UN agenda, and beyond that, some of the most urgent demands of human societies. Understanding the role of DL in EO for extracting nationwide geospatial statistical data has far-reaching societal implications for policy development and decision making. Going beyond the SDG agenda, DL and EO can play a significant role in other international agendas such as the New Urban Agenda (NUA) (https://habitat3.org/the-new-urban-agenda) or the Sendai framework for disaster risk reduction.




%





\ifCLASSOPTIONcaptionsoff
  \newpage
\fi



%

%
\bibliographystyle{IEEEtran}
\bibliography{references,ref_mendeley,SDGs_devis,SDGs_jan,SDGs_gustau,SDGs_ronny,tackling_climate_change_with_ml_NEW}


%








\end{document}